\pdfoutput=1
\documentclass[11pt]{article}

\usepackage[final]{acl}

\usepackage{times}
\usepackage{latexsym}

\usepackage[T1]{fontenc}
\usepackage[utf8]{inputenc}
\usepackage{amssymb} 

\usepackage{microtype}

\usepackage{inconsolata}

\usepackage{graphicx}
\usepackage{amsmath}
\usepackage{multirow}
\usepackage{booktabs}
\title{Better Generalizing to Unseen Concepts: An Evaluation Framework and An LLM-Based Auto-Labeled Pipeline for Biomedical Concept Recognition}
\author{
 \textbf{Shanshan Liu\textsuperscript{1,2}},
 \textbf{Noriki Nishida\textsuperscript{1}},
 \textbf{Fei Cheng\textsuperscript{3}},
 \textbf{Narumi Tokunaga\textsuperscript{1}},
 \textbf{Rumana Ferdous Munne\textsuperscript{1}},
\\
 \textbf{Yuki Yamagata\textsuperscript{4, 5}},
 \textbf{Kouji Kozaki\textsuperscript{6}},
 \textbf{Takehito Utsuro\textsuperscript{2}},
 \textbf{Yuji Matsumoto\textsuperscript{1}},
 \\
    \textsuperscript{1}RIKEN AIP \ \ \ 
    \textsuperscript{2}University of Tsukuba \ \ \
    \textsuperscript{3}Kyoto University \ \ \
    \textsuperscript{4}RIKEN R-IH \ \ \ \\
    \textsuperscript{5}RIKEN BRC \ \ \ 
    \textsuperscript{6}Osaka Electro-Communication University\\
    \{shanshan.liu, noriki.nishida, narumi.tokunaga, rumanaferdous.munne, \\
    yuki.yamagata, yuji.matsumoto\}@riken.jp\\
    feicheng@i.kyoto-u.ac.jp; 
    kozaki@osakac.ac.jp; utsuro@iit.tsukuba.ac.jp\\
}
\begin{document}
\maketitle
\begin{abstract}

% Ontology-based biomedical concept recognition (CR) maps free-text passages to ontology concepts, enabling structured knowledge extraction. However, existing methods struggle with limited training data, complex fine-grained ontology, and implicitly expressed concepts.
% We address these challenges via two complementary strategies: (1) an LLM-based auto-labeling pipeline for large-scale annotation, and (2) three types of hierarchical indices for structured prediction.
% Our pipeline labels passages with concepts from the Human Phenotype Ontology (HPO) and the Homeostasis Imbalance Process (HoIP) Ontology, producing synthetic training data that improves generalization to unseen concepts. We further propose hybrid indexing schemes that combine semantic and ontological signals, yielding strong gains in recognition accuracy and candidate-space reduction. We analyze the effectiveness of our strategies, including overall performance and strictly unseen concept recognition.
% We prove our methods offer a scalable framework for robust, ontology-aware CR in low-resource biomedical settings. Code and data: \url{https://github.com/bio-ie-tool/hi-ald}.

% Mention-agnostic Biomedical Concept Recognition (MA-BCR) aims to recognize ontology concepts directly from text, which is particularly suited for handling implicit concept expressions. However, the scarcity of human-annotated data means that most concepts remain unseen during recognizer development, presenting a critical challenge for generalization.

Generalization to unseen concepts is a central challenge due to the scarcity of human annotations in Mention-agnostic Biomedical Concept Recognition (MA-BCR). This work makes two key contributions to systematically address this issue. First, we propose an evaluation framework built on hierarchical concept indices and novel metrics to measure generalization. Second, we explore LLM-based Auto-Labeled Data (ALD) as a scalable resource, creating a task-specific pipeline for its generation. Our research unequivocally shows that while LLM-generated ALD cannot fully substitute for manual annotations, it is a valuable resource for improving generalization, successfully providing models with the broader coverage and structural knowledge needed to approach recognizing unseen concepts.
Code and datasets are available at \url{https://github.com/bio-ie-tool/hi-ald}.

%However, existing methods are under-evaluated on their ability to generalize to unseen concepts, and remain unexplored when trained on LLM-based auto-labeled data (ALD).

%We build a task-specific auto-labeling pipeline to construct large ALDs for Human Phenotype Abnormality and Homeostasis Imbalance Process concepts. 
%For analysis of whether the model learns hierarchical information by ALD, we introduce three hierarchical indices (OSI, SSI, OSSI) with two metrics: U-RC (hierarchical closeness to unseen gold) and U-CS (unseen candidate set size).  
%Compared with limited manually labeled data, training with ALD lowers F1 but consistently improves generalization ability (↑U-RC, ↓U-CS). These results position ALD as complementary coverage and structural signal, not a replacement for manual labels on mention-agnostic CR.

\end{abstract}

\section{Introduction}

Biomedical Concept Recognition (BCR) —identifying ontology concepts expressed in free-text passages—is foundational for knowledge-intensive applications. Accurate and efficient CR systems may facilitate the construction and maintenance of structured biomedical knowledge bases, accelerate knowledge discovery, and ultimately support downstream applications such as therapeutic innovation.
 % (e.g., by enabling faster retrieval and surfacing novel concepts)

Most prior works formulate the BCR task as mention-based recognition: detect a text span (a ``mention'') in the text and then link it to an ontology concept \cite{DBLP:journals/biodb/LiSJSWLDMWL16,luo2021phenotagger, wang2023exploringincontextlearningability, 10.1093/bioinformatics/btae104, groza2024evaluation}. While being effective when concepts are explicitly expressed, this setting is less aligned with biomedical discourse, where many concepts are expressed implicitly. This misalignment persists even when mentions are annotated.

For instance, in an abstract included by the HPO GSC+ corpus \cite{https://doi.org/10.1155/2017/8565739}, the span \textit{``melanin pigment synthesis in the hair, skin, and eyes''} is annotated as \textit{Generalized hypopigmentation (HP:0007513)}. The concept is implicit with respect to that span—it is neither the ontology label nor a semantically equivalent paraphrase—and the annotation is supported by intra-abstract cues (e.g., hypopigmentation, reduced retinal pigment) rather than surface matching. %The more semantically complex a concept is, the less likely it is to appear verbatim or via surface-level paraphrases. 

In addition, mention-based supervision requires costly two-level annotation (mention spans and mention-concept mappings), limiting scalability in domains where expert labeling is expensive. We therefore study \textbf{MA-BCR}, which directly identifies ontology concepts from a passage without requiring intermediate mention spans. This formulation better matches biomedical discourse and alleviates annotation burden (no need for mention spans for model training and inference).

Independent of the task formulation, a key requirement for real-world deployment of recognizers is the ability to \textbf{generalize to unseen concepts}, given that manually labeled datasets (MLDs) cover only a small fraction of concepts in biomedical ontologies due to the high requirements on expertise and annotation time. For example, the HPO GSC+ corpus comprises 228 abstracts and covers approximately 2.4\% of Human Phenotype Abnormality (HPA) concepts in the Human Phenotype Ontology (HPO) \cite{10.1093/nar/gkad1005}. Despite its importance, this capability has been under-evaluated. 
Beyond results showing that recognizing unseen concepts was nearly infeasible for a recognizer MA-COIR \cite{liu2025macoirleveragingsemanticsearch}, \textbf{clear experimental evidence on unseen-concept recognition remains scarce}.

%To systematically probe model generalization in recognizing unseen concepts—where models cannot predict the target directly—we propose an innovative evaluation framework. This framework is built upon two core components. First, we develop three types of hierarchical indices (ontology-aware, semantic-based, and hybrid) which establish a structured space to make generalization measurable.
%that serve as the structural basis for measuring generalization. 

To systematically probe model generalization in recognizing unseen concepts—where models cannot predict the target directly—we propose an innovative evaluation framework. This framework is built upon two core components. First, we develop three types of hierarchical indices (ontology-aware, semantic-based, and hybrid), while each of them can serve as a structured concept space to make generalization measurable. Second, we introduce two new metrics that operate within this space: Unseen Recall-oriented Closeness (U-RC), which quantifies how close a model gets to the unseen concept, and Unseen Candidate set Size (U-CS), which measures how much the model shrinks the search space compared to the full label sets.

%Unseen Recall-oriented Closeness (U-RC), which measures how close a model’s prediction is to the unseen gold concept in the hierarchy, and Unseen Candidate set Size (U-CS), which measures how much the model shrinks the search space for an unseen concept compared to the full label sets.

Our evaluation framework provides the tools to measure generalization, yet the question of how to improve it remains. A practical pathway is to expand concept coverage and contextual diversity through Auto-Labeled Data (ALD). %Insights from biomedical entity normalization suggests that representations aligned across paraphrases/synonyms generalize beyond surface forms (e.g., SapBERT \cite{liu-etal-2021-self}; BioSyn \cite{sung-etal-2020-biomedical}). 
Given their success as automatic annotation providers in other NLP tasks \cite{tan-etal-2024-large}, Large Language Models (LLMs) present a scalable avenue for higher concept coverage. This potential is especially important for MA-BCR, a task that requires good language understanding capabilities for assigning passage-level labels without relying on surface mentions. We therefore design a novel auto-labeling pipeline based on LLMs. Its components are tailored for MA-BCR, and the final design is determined through extensive empirical validation. We employ our evaluation framework to assess this data-centric approach, specifically asking:
\begin{itemize}
    \item \textit{RQ1: Is LLM-generated ALD sufficient for training a recognizer that performs effectively on both seen and unseen concepts?} 
    \item \textit{RQ2: Despite the noise, does ALD improve generalization to unseen concepts?}
\end{itemize}

Our investigation proceeds as follows. We first quantify the quality of our ALD by benchmarking it against Manual-Labeled Data (MLD). We then employ the MA-COIR recognizer, which outputs pre-defined hierarchical concept indices as predictions, allowing for a grounded assessment of generalization. Finally, by training MA-COIR on different scales of MLD and ALD, we compare their effectiveness in enhancing model performance.

On two pairs of concepts and ontologies, HPA concepts in HPO and Homeostasis Imbalance Process (HoIP) concepts in HoIP Ontology \cite{Yamagata2024}, models trained on ALD consistently achieve substantial gains in unseen-oriented metrics as the data volume increases, despite exhibiting lower exact-matching accuracy than their MLD-trained counterparts. This indicates that exposure to broad and noisy supervision enhances hierarchy-aware generalization even when exact-match accuracy lags. We position LLM-based auto-labeling not as a replacement for manually labeling, but as a complementary source of coverage and structural signal for unseen concept recognition. Our contributions are as follows:
\begin{itemize}
    % \item We establish a controlled unseen-aware evaluation setup for mention-agnostic CR, aligning assessment with real deployment needs.
    \item We introduce an LLM-based auto-labeling pipeline for MA-BCR, and construct two large auto-labeled datasets. To our knowledge, LLM-based automatic labeling has not been explored in MA-BCR before our work.
    \item We design three types of hierarchical indices and two metrics for directly assessing hierarchy-aware learning and search-space reduction on unseen concepts.
    \item Results show that models trained on large-scale ALD underperform MLD-trained models on F1 but generalize better to unseen concepts, clarifying ALD’s role in MA-BCR.
\end{itemize}

% We hope this work drives new efforts in scalable concept recognition and structured biomedical knowledge discovery.

\begin{figure*}[t]
\centering
  \includegraphics[width=0.94\linewidth]{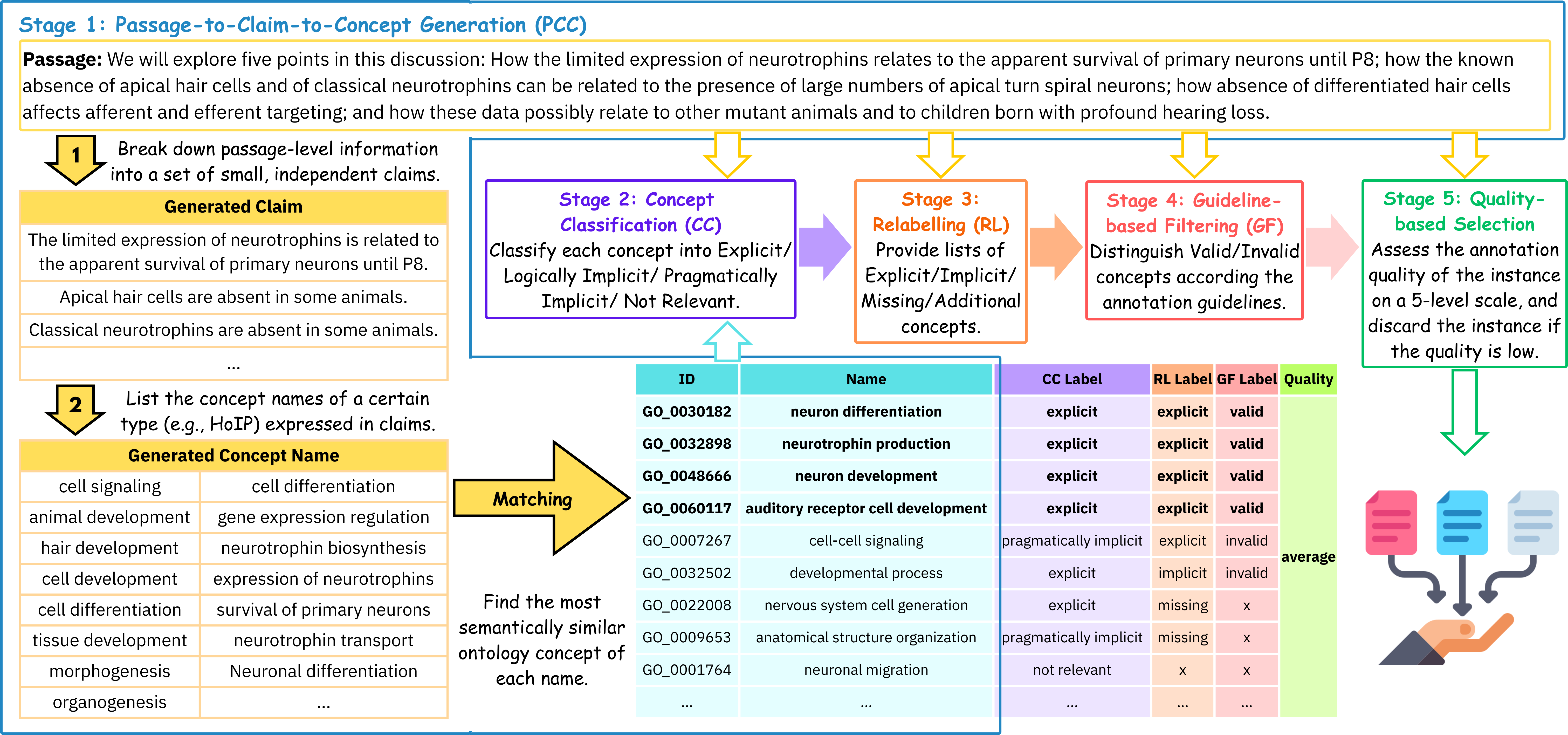}
  \caption{Overview of the LLM-based Auto-labeling Pipeline. Given an input passage, the pipeline begins by generating intermediate claims (Arrow 1), followed by generating candidate concept names from the claims (Arrow 2). The resulting names are matched to ontology terms by comparing the representations (encoded by SapBERT) of generated and ontological names, forming a preliminary list of concept candidates, shown in the blue-shaded table. Then concept classification, relabeling, and guideline-based filtering steps are applied to get the final annotations (highlighted in bold). The instance will be taken as a training instance if its quality meets the requirement.}
  \label{fig:alp}
\end{figure*}

\section{Related Works}
\paragraph{Biomedical Concept Recognition.}
Biomedical concept recognition has traditionally been approached as a two-stage task: mention detection followed by entity linking. Early systems leverage neural architectures for named entity recognition and linking \cite{DBLP:journals/biodb/LiSJSWLDMWL16,luo2021phenotagger}, while more recent methods introduce LLM-based pipeline \cite{wang2023exploringincontextlearningability, 10.1093/bioinformatics/btae104, groza2024evaluation, el-khettari-etal-2024-mention} or reformat that task as a one-step sequence-to-sequence generation framework \cite{liu2025macoirleveragingsemanticsearch}. However, these approaches often struggle with concept ambiguity, limited coverage, and implicit expressions—issues exacerbated by the sparsity of high-quality annotated data.

\paragraph{Hierarchical Indexing.}
Hierarchical indexing has proven effective in tasks involving large output spaces, such as extreme multi-label classification \cite{zhang2021fast, kharbanda2022cascadexml}, document retrieval \cite{tay2022transformermemorydifferentiablesearch}, and biomedical concept recognition (BCR) \cite{liu2025macoirleveragingsemanticsearch}. By organizing labels or documents into trees based on semantic information through K-Means clustering, these methods improve efficiency and precision. Despite its promise, hierarchical indexing remains under-explored in BCR, where ontologies naturally provide structured concept taxonomies.

% \paragraph{Synthetic Data.}
% Synthetic data have gained attraction in low-resource NLP tasks. Recent work has demonstrated the utility of LLM-generated training data for tasks like information extraction, question answering, and text classification \cite{li2023synthetic, josifoskietal2023exploiting, longetal2024llms, li2024datagenerationusinglarge}. However, few studies have examined the construction of large-scale synthetic datasets for biomedical CR.

\section{Methodology}

\subsection{Task formulation}

% In biomedical texts, concepts may be involved in a passage without being explicitly mentioned—either as implicit premises or consequences required by the passage’s logic, or as plausible inferences based on domain knowledge. We cherish these implicit concepts, so that the CR task is not only to recognize explicit expressed concepts but also implicit ones.

Mention-agnostic Biomedical Concept Recognition (MA-BCR) aims to directly identify a subset of ontology-defined concepts \(\{C'_1, ..., C'_p\} \subseteq O\) that are referenced in a given text \(Q\), while the ontology \(O = \{C_1, ..., C_n\}\) is constructed by domain experts.
We frame this as an end-to-end generative task: the model directly produces a sequence of unique concept identifiers 
\(\{I_{C'_1}, ..., I_{C'_p}\}\) for \(Q\), with each \(I_{C_i}\) corresponding to a concept \(C_i\) in \(O\).

MA-BCR diverges fundamentally from conventional NER+EL pipelines for BCR. %which rely on explicit mentions as intermediates. 
It is specifically designed to capture not only \textbf{explicit} concepts—those directly realized in the text via surface forms or synonyms—but also \textbf{implicit} concepts. The latter are referenced through logical implications, domain-specific inferences, or unspoken premises, making them unlikely to appear verbatim or as simple paraphrases.

%Explicit concepts are directly realized in the text via surface forms or synonyms. In contrast, implicit concepts are referenced through logical implications, domain-specific inferences, or unspoken premises, and are less likely to appear verbatim or via surface-level paraphrases.

%For example, an abstract in the HPO GSC+ corpus states: \textit{``...important for melanin pigment synthesis in the hair, skin, and eyes... Hair color ranged from light blond to brown. Skin was type I in 3 and type II in 3... reduced retinal pigment in 5.''} This implicitly expresses the phenotypes \textit{Generalized hypopigmentation (HP:0007513)} and \textit{Hypopigmentation of the fundus (HP:0007894)}. While such concepts were annotated in the corpora, they remain implicit, as the annotated mentions of them—\textit{``melanin pigment synthesis in the hair, skin, and eyes''/``reduced retinal pigment''}—do not match ontology labels or synonyms directly. Recognizing such concepts requires integrating anatomical knowledge (e.g., the relationship between the retina and the fundus) and domain-specific reasoning (e.g., ``melanin pigment synthesis in hair, skin, and eyes'' is a statement of pathology, not a symptom or phenotype, but it indirectly suggests that the patient will have generalized hypopigmentation).

% \subsection{Our Approach}

% To address two persistent bottlenecks in biomedical CR-limited annotation resources and the scale of ontology-defined concept spaces, we propose an LLM-based auto-labeling pipeline and three types of hierarchical indices.

\begin{figure*}[t]
\small
\centering
    \includegraphics[width=0.97\linewidth]{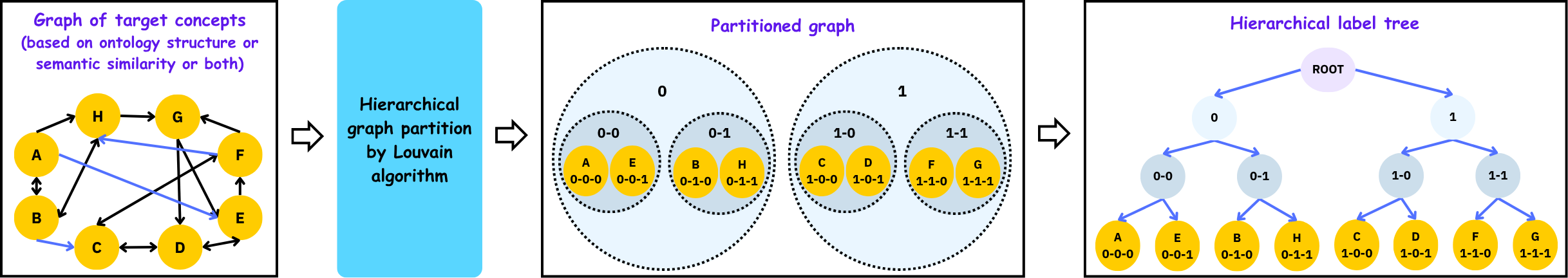} 
    \caption{Overview of hierarchical search index construction. Ontology concepts are represented as nodes in a graph (yellow), where edges reflect either ontological relations (blue), semantic similarities (black), or both. A graph partitioning algorithm (e.g., Louvain) is then applied to recursively divide the graph into nested subgraphs. In the illustrated example, the initial graph is first partitioned into two coarse-level clusters (labeled 0 and 1), which are further subdivided into finer clusters (e.g., 0-0, 0-1, 1-0, 1-1). Based on the partitioned structure, a hierarchical label tree is constructed: internal nodes (gray) represent clusters, while leaf nodes (yellow) correspond to individual ontology concepts. Each concept is assigned an index reflecting its position in the tree (e.g., concept A is labeled 0-0-0, indicating its membership in cluster 0, then 0-0, and finally its position).}
    \label{fig:graph-index}
\end{figure*}

\subsection{LLM-based Auto-Labeling Pipeline}
\label{sec:alp}

LLMs offer a powerful foundation for automatic biomedical concept annotation. Trained on vast textual corpora, LLMs encode extensive biomedical knowledge and can infer relevant concepts even when they are not explicitly mentioned.
%—a crucial capability in biomedical texts, where implicit expression is common. 
Furthermore, LLMs are highly adaptable: with prompt-based control, a single model can function as a generator, classifier, filter, or evaluator, enabling the construction of a flexible, modular pipeline.

To figure out \textbf{what label quality can LLM-based auto-labeling achieve for MA-BCR}, we carefully design an LLM-based pipeline with five stages (Figure~\ref{fig:alp}). Each stage is described in the following paragraphs. Full implementation details and inference costs are provided in Appendix~\ref{app:alp}.

%We propose an LLM-based pipeline for constructing auto-labeled datasets at scale. As shown in Figure~\ref{fig:alp}, our pipeline consists of five stages. We introduce each stage as following. (1) Given an input passage, the pipeline begins by generating intermediate claims, followed by candidate concept names derived from these claims. The resulting names are semantically matched to ontology terms via a k-Nearest Neighbor (kNN) search, forming a preliminary list of concept candidates. (2) A concept classification step then filters out irrelevant candidates by categorizing them into four predefined types. (3) The relabeling module refines this list by identifying missing concepts (mentioned in the passage but not included in the list) and removing unsupported ones. (4) Next, guideline-based filtering applies domain-specific annotation criteria to further eliminate invalid entries. (5) In the final step, we assess the quality of passage-annotations, and select high-quality samples as instances. Full implementation details and design rationale are provided in Appendix~\ref{app:alp}.

\paragraph{Stage 1: Passage-to-Claim-to-Concept Generation.}
\label{sec:pcc}

Direct generation of concept names from passages often yields low recall and limited relevance, particularly for complex biomedical concepts (e.g., $\sim$30\% recall on HoIP-MLD). To address this, we introduce an intermediate \textit{claim generation} step, in which key biomedical assertions are summarized into concise claims (Figure~\ref{fig:alp}-Arrow 1). These claims serve as more focused inputs for concept derivation, substantially improving both coverage and grounding. On HoIP-MLD, this step raises recall from $\sim$30\% to $\sim$62\%.

Concept names generated from claims are subsequently mapped to ontology labels via semantic matching (Figure~\ref{fig:alp}-Arrow 2). We encode both the generated names and all ontology concept names (including synonyms) into 768-dimensional vectors using mean-pooled SapBERT embeddings.\footnote{Model card: cambridgeltl/SapBERT-from-PubMedBERT-fulltext}
Candidate matches are retrieved with Faiss \cite{douze2024faiss}, using L2 distance as implemented by default. We select the top-1 match for each generated name if the similarity score is $\geq$ 0.6, forming a preliminary candidate list (illustrated in the blue-shaded table in Figure~\ref{fig:alp}).

\paragraph{Stage 2: Concept Classification.}
%The preliminary candidate concepts obtained after Passage-to-Claim-to-Concept generation often contain both relevant and irrelevant entries. 

The preliminary candidate list often contains both relevant and irrelevant concepts.
This arises for two reasons: (i) LLM may generate irrelevant concepts due to its limited capability, and (ii) similarity-based matching may link a name to semantically close but irrelevant terms. %(e.g., a generated name ``AA'' matched to ontology terms ``AAA'' or ``AAB''). 
To resolve this, we formulate a filtering process as a four-way classification problem. Each candidate is categorized as: (1) \textit{Explicit}, (2) \textit{Logically Implicit}: concepts required by the passage’s logical structure, e.g., necessary premises, (3) \textit{Pragmatically Implicit}: concepts plausibly inferred from domain knowledge or text content, though not logically required, or (4) \textit{Not Relevant}. % or consequences

This finer-grained categorization serves two purposes. First, it improves filtering accuracy: preliminary experiments show that this four-class classification outperforms both binary (relevant/not relevant) and three-class (explicit/implicit/not relevant) settings in distinguishing relevant concepts. Second, it enables flexible downstream usage: applications prioritizing precision may retain only explicit and logically implicit concepts, while discovery-oriented tasks can also benefit from pragmatically implicit ones. In our experiments, concepts classified as ``Not Relevant'' are discarded.

%(e.g., clinical decision support)

%To reduce false positives from similarity-based linking, we classify each candidate into four categories: (1) Explicit, (2) Logically Implicit Concept: Implicit concepts required by the passage’s logical structure (e.g., necessary premises or consequences),
%(3) Pragmatically Implicit Concept: Plausibly inferred based on domain knowledge or text content, though not logically necessary, or (4) Not Relevant. This finer-grained classification allows the model to better distinguish relevant concepts, particularly those not overtly mentioned. It also supports flexible downstream applications: when precision is critical (e.g., clinical decision support), only explicit and logically implicit concepts may be retained; for discovery-oriented tasks, pragmatically implicit concepts can provide valuable insights. %(e.g., knowledge graph expansion), 
%This definition aims to improve the flexibility and accuracy of labeling and capture such implicit concepts is critical for building comprehensive biomedical knowledge graphs and ontological resources. When only high-confidence knowledge is required, passing only explicit or explicit and logically implicit concepts to the next stage is appropriate. In contrast, for knowledge discovery tasks, including pragmatically implicit concepts may facilitate the identification of novel insights.

\paragraph{Stage 3: Relabeling.}
Even after Concept Classification, candidate concepts can still include false positives and false negatives. Residual false positives mainly stem from misclassifications in the Concept Classification stage, whereas residual false negatives arise from omissions during the Passage-to-Claim-to-Concept generation or from overly aggressive filtering in Concept Classification.

To address these issues, we employ an LLM to evaluate each passage–concept pair. For each sample, the LLM generates a justification highlighting strengths (concepts present) and weaknesses (concepts missing), and categorizes the candidate as explicit, implicit, or missing. It also identifies additional relevant concepts mentioned in the passage but absent from the candidate list. This feedback is then used to add missing concepts and remove incorrect ones, producing a more accurate and comprehensive set of annotations. 
%The LLM also provides a quality rating (very poor to excellent) for each passage–concept pair, supporting downstream selection of high-quality training instances.

%We further refine annotations by prompting the LLM to reassess each passage–concepts pair: it identifies missing concepts, removes incorrect ones, and justifies its assessment. Concepts are re-tagged as explicit, implicit, or irrelevant, improving both coverage and precision.

\paragraph{Stage 4: Guideline-based Filtering.}

Automatically annotated concepts may not fully align with the standards established by manual annotation. To ensure consistency and compliance with official criteria, we leverage existing annotation guidelines. 

For HOIP concepts, the CRAFT corpus provides a 47-page guideline with over 100 examples and 50 figures. We refined the guideline by removing figures and reorganizing textual content, then used an LLM to generate a concise summary (2,128 characters). This summary is incorporated into the prompt that directs the LLM to review each passage and its candidate concept list, classifying each concept as valid or invalid according to the summarized guideline.

\begin{table*}[t]
\small
% \footnotesize
    \centering
    \begin{tabular}{c|cc|ccc|ccc}
    \toprule
    \multirow{2}{*}{\textbf{Case}}& \multicolumn{2}{c|}{\textbf{Ontology}}& \multicolumn{3}{c|}{\textbf{MLD}}&\multicolumn{3}{c}{\textbf{ALD}} \\
& \textbf{Uni.Con} & \textbf{Name} & \textbf{Passage} & \textbf{Concept} & \textbf{Uni.Con (Coverage)}  & \textbf{Passage} & \textbf{Concept} &\textbf{Uni.Con (Coverage)}\\
\midrule
HPA & 18,354 & 40,341 & 228 & 1,423 & 431 (2.35\%) & 54,301  & 197,824 & 12,725 (69.33\%)\\
HoIP & 29,367 & 117,072 & 3,621 & 7,855 & 690 (2.35\%) & 34,097 & 370,672 & 15,976 (54.40\%) \\
\bottomrule
    \end{tabular}
    \caption{Dataset Statistics. Manually labeled data (MLD) are sourced from HPO-GSC+ and CRAFT; automatically labeled data (ALD) are generated via our auto-labeling pipeline. ``Uni.Con'' denotes the number of unique concepts. ``Coverage'' refers to the percentage of ontology concepts presented in the dataset.}
    \label{tab:data-stats}
\end{table*}

\paragraph{Stage 5: Quality-based Selection.}
\label{sec:qs}

Even after relabeling, annotation quality can vary across passages, and low-quality samples (passage-concept pairs) risk introducing noise into downstream applications. To mitigate this, we implement a quality-based selection step that filters at the passage level.

In this stage, each sample is evaluated by an LLM according to a five-tier rating scheme: (1) \textit{Very Poor}: majority of concepts missing or irrelevant, (2) \textit{Poor}: partial concept inclusion but lacking relevance, (3) \textit{Average}: Most concepts present with basic contextualization, (4) \textit{Good}: Comprehensive concept coverage with coherent context, (5) \textit{Excellent}: full concept integration with clear explanations. In our experiments, only samples rated ``Average'' or above are retained.

Unlike earlier stages, which operate at the concept level, this stage evaluates the overall reliability of each sample. Passages dominated by irrelevant or incomplete annotations are discarded, ensuring dataset-level consistency, reducing the risk of noisy samples propagating into downstream applications.

% \paragraph{Summary.}
% Overall, errors introduced by LLM generation and similarity-based matching can propagate across stages, and later filtering and relabeling only partially mitigate them. As a result, residual noise and passage-level quality variation remain, motivating systematic evaluation of auto-labeled data quality and downstream usefulness.

\subsection{Hierarchical Index Design}
\label{sec:hi}

Traditional indexing in large label spaces relies on semantic similarity alone. In BCR, however, structured ontologies provide a second signal—explicit parent-child relations encoding curated expert knowledge. We explore both signals to construct more meaningful and scalable indices.

Following existing practices \cite{liu2025macoirleveragingsemanticsearch}, we enforce a hierarchical structure of label tree where each concept is a leaf node (as shown in Figure~\ref{fig:graph-index}). Internal nodes are limited to at most 10 children, striking a balance between decomposition granularity and model tractability. To handle both tree-like (HPO) and graph-like (HoIP) ontologies, we construct indices using graph-based methods.

% first convert the ontology into a weighted graph, then derive the hierarchy via recursive partitioning.

We experiment with three graph construction strategies, resulting in three types of indices: 

(1) \textbf{OSI (Ontology-aware Search Index)}: Edges follow ontology parent-child links with fixed weights of 1.0. 

(2) \textbf{SSI (Semantic Search Index)}: Edges connect top-10 nearest embedding neighbors; weights reflect semantic similarity range from $0.0 \text{-} 1.0$.\footnote{Weights are calculated as the same way for generated name-ontology concept matching in Section~\ref{sec:pcc}, Stage 1.} 

(3) \textbf{OSSI (Ontology-Semantic Search Index)}: A hybrid combining OSI and SSI, prioritizing ontology structure while integrating semantic proximity.

We then do hierarchical graph partitioning on the graph of concepts, which yields trees that encode ontology structure or preserve semantic coherence.\footnote{Details are in Appendix~\ref{app:gp-algo}.}  \textbf{Corresponding indices help us to understand what type of hierarchical information has been learned by the recognizer.}
Comparing SSI with ssID provided by \citet{liu2025macoirleveragingsemanticsearch}, the biggest difference is that we use graph partitioning for clustering, while they applied K-means method.

\section{Experiment Design}

\subsection{Datasets}
% \paragraph{Manually labeled datasets (MLDs).}
% We evaluate our method on two manually annotated corpora covering distinct biomedical concept types: (1) the HPO GSC+ corpus, annotated with Human Phenotypic Abnormality (HPA) concepts, and (2) a modified version of the CRAFT corpus, repurposed for Homeostasis Imbalance Process (HoIP) concepts.

% The scope of HPA concepts, are descendants of the ontology class \texttt{HP:0000118 - Phenotypic Abnormality} of HPO, while of HoIP concepts, are Homeostasis Imbalance Process concepts included in the HoIP ontology.  %These constraints ensure conceptual consistency and alignment with our target ontologies.

% Dataset statistics are summarized in Table~\ref{tab:data-stats}. Details of MLD preprocessing for our task is in Appendix 1.

\paragraph{Target Concepts.}
Human Phenotypic Abnormality (HPA) concepts are defined as descendants of \texttt{HP:0000118-Phenotypic Abnormality} in HPO. Homeostasis Imbalance Process (HoIP) concepts correspond to Homeostasis Imbalance Process terms included in the HoIP Ontology.

\paragraph{Manual-Labeled Datasets (MLDs).}
We evaluate on two manually annotated corpora: (1) HPA-MLD, from the HPO GSC+ corpus, and (2) HoIP-MLD, adapted from the CRAFT corpus. Statistics are in Table~\ref{tab:data-stats}. Examples are in Appendix~\ref{app:dataset}.

\paragraph{Auto-Labeled datasets (ALDs).}
To evaluate the scalability and effectiveness of our auto-labeling pipeline, we construct two large-scale ALDs for HPA and HoIP concepts, respectively. 
For each target concept, we retrieve up to 10 recent PubMed abstracts using the concept’s preferred label as a query, via the NIH E-utilities API.\footnote{\url{https://www.ncbi.nlm.nih.gov/books/NBK25500/}} Duplicates are removed to ensure passage uniqueness. These passages are then annotated using our auto-labeling pipeline to produce two ALDs for downstream evaluation. Dataset statistics are reported in Table~\ref{tab:data-stats}.

\paragraph{Data Splits.}
% We design evaluation splits to support two key analyses: (i) the effectiveness of auto-labeling pipeline, and (ii) the overall data quality of ALD compared to manually labeled data (Table~\ref{tab:data-quality}).

For auto-labeling evaluation (Table~\ref{tab:ald-ori-hpo}), we apply the ALD pipeline to 228 HPA-MLD and a randomly sampled 500-instance subset of HoIP-MLD. Manual labels are used as gold standard to evaluate how each module impacts quality.

For recognizer training and evaluation (Table~\ref{tab:HoIP-ALD-res}), we split the data as follows. For \textbf{HPA}, we use 45 MLD instances (20\%) for development and 183 for testing. Due to the small size of HPA-MLD, we train models only on HPA-ALD (max to 47,152 instances). 
For \textbf{HoIP}, we adopt the CRAFT dev split (375 instances), and evaluate in two training setups:  
  (1) training on HoIP-MLD (max to 2,458 instances) and testing on 375 sampled MLD instances (MLD$\rightarrow$MLD);  
  (2) training on HoIP-ALD (max to 16,415 instances) and testing on 3,246 held-out HoIP-MLD instances (ALD$\rightarrow$MLD).  

To assess generalization, we ensure that at least 30\% of test concepts are unseen during training. This is achieved by filtering out training passages containing those concepts.  We vary the training size $M = 200 \cdot 2^k,\; k=0,\dots,7.$ to test performance scalability. To maintain stable model selection, we always include 100 core training passages with concept distributions closest to the dev set. Additional $(M-100)$ training instances are sampled randomly. Statistics are provided in Table~\ref{tab:split-stats}.
% $M\in\{200, 400, 800, 1600, 3200, 6400, 12800, 25600\}$

% \paragraph{Data Splits.}
% We design evaluation splits to test both standard performance and generalization to unseen concepts.
% For HPA, 45 (20\%) are used for development, and the remaining 183 for testing models trained on the ALD. Due to the limited size, we do not train any model directly on HPA MLD. For HoIP, we follow the original CRAFT dev split (375 instances) and construct two test sets: (1) a 375-instance subset for evaluating MLD-trained models, and (2) the full 3,246-instance set for evaluating ALD-trained models.

% To simulate unseen concept scenarios, we ensure that at least 30\% of concepts in each test set do not appear in training. This is achieved by selecting 30\% of test concepts and excluding any training instance that mentions them. To maintain stable model selection, we always include 100 core training instances with concept distributions closely matching the dev set. The remaining $(M - 100)$ instances are randomly sampled to increase coverage. Models are trained on progressively larger subsets of the ALD: \(M \in \{200, 400, 800, 1600, 3200, 6400, 12800, 25600\}\). We also include full-pool training runs: 47,152 instances for HPA ALD, and 2,458 (MLD) or 16,415 (ALD) instances for HoIP. Detailed concept coverage statistics are provided in Appendix~\ref{app:dataset}.

\subsection{Models}

\paragraph{Models for ALD construction.}
To keep the auto-labeling pipeline scalable and accessible, we constrain LLMs to either (i) models runnable on a single consumer-grade GPU or (ii) cost-effective commercial APIs. After extensive testing across GPT and LLaMA variants, we adopt LLaMA-3-8B\footnote{Model card: meta-llama/Meta-Llama-3-8B-Instruct} for claim and concept generation due to its superior fluency, specificity, and domain grounding in both HPA and HoIP terms. For downstream tasks such as classification, relabeling, and filtering, we use GPT-4o-mini, balancing high reasoning ability with cost efficiency. This setup ensures useful annotations without sacrificing accessibility. More details are in Appendix~\ref{app:alp}.

\paragraph{Concept recognizer.}
We adopt MA-COIR \cite{liu2025macoirleveragingsemanticsearch}—a BART-based seq2seq recognizer that outputs ontology concept indices from free text—as our canonical mention-agnostic model. This choice is methodological: an index-producing recognizer lets us (i) make hierarchical structure explicit to learn and (ii) quantify unseen behavior with hierarchy-aware metrics easily.

%To isolate the role of index structure, 
We keep the architecture and training recipe\footnote{Hyperparameters are listed in Appendix~\ref{app:model}.} fixed and swap MA-COIR’s original semantic index for three hierarchical variants: an \textbf{ontology-structured} index (OSI), a \textbf{semantic-structured} index (SSI), and a \textbf{hybrid ontology–semantic} index (OSSI). This design enables a clean assessment of whether structural signals affect generalization.

Our goal is not to propose a new model architecture nor to compete for state-of-the-art. Cross-architecture baselines (often mention-based) are incapable of learning with a designed hierarchy and are orthogonal to our research questions.
%it is to measure the generalization ability under different training data and index designs. Cross-architecture baselines (often mention-based) are incapable of this analysis and are orthogonal to our questions. 
Prior work has already positioned MA-COIR as a strong MA-BCR approach; our SSI follows a similar design and serves as a strong baseline. Accordingly, we \textbf{do not report} comparisons to other models, nor do we re-run the original MA-COIR indexing. %instead, all comparisons are within the same model class to disentangle the effects of index structure and training data (MLD vs.\ ALD).

\subsection{Evaluation Metrics}

We evaluate from three complementary views: (i) exact-match accuracy, (ii) hierarchy-aware closeness to unseen concepts, and (iii) search space of unseen concepts inferable by predictions. 

For a passage $p$, let $Y(p)$ be its gold concept set and $\hat{Y}(p)$ be the set produced by the recognizer. Each concept $x$ is represented by an index sequence $I_x = [i_1,\dots,i_{|I_x|}]$. Let $\mathrm{lcp}(I_a, I_b)$ be the length of their longest common prefix.

\paragraph{Exact Match.}
We report precision, recall, and micro-F1. A prediction is correct if it exactly matches any gold concept in $Y(p)$.

\paragraph{Unseen Recall-oriented Closeness (U-RC).}
We compute closeness for each gold concept unseen during training and then average across all of them as the U-RC.  
Let $\mathcal{P}_{\text{test}}$ be the test passages. For a passage $p\in\mathcal{P}_{\text{test}}$, let
$G_{\text{us}}(p)=\{\,g\in Y(p)\mid g\ \text{is unseen}\,\}$ and $\hat{Y}(p)$ be the model's predictions for $p$. The U-RC is calculated as:
\[
\mathrm{U\text{-}RC}
=\frac{\sum_{p}\;\sum_{g\in G_{\text{us}}(p)}
\displaystyle \max_{\hat{y}\in \hat{Y}(p)}
\frac{\mathrm{lcp}(I_g, I_{\hat{y}})}{|I_g|}}
{\sum_{p} |G_{\text{us}}(p)|}.
\]
If $\hat{Y}(p)=\varnothing$, the max is defined as $0$.
Higher is better; it reflects how closely in the hierarchy, predictions approach the correct \emph{unseen} concepts.

\paragraph{Unseen Candidate-set Size (U-CS).}
To evaluate the model's utility in reducing the search space for unseen concepts, we propose the U-CS. 

% U-RC measures hierarchical proximity but not the reduction in search space. A prediction and the gold concept may reside in the same penultimate cluster that contains 2 or more candidates—a critical difference U-RC misses. Therefore, we propose the U-CS.

%, which directly quantifies how much the model shrinks the search space for an unseen concept relative to the full label set.

For each unseen gold $g\in G_{\text{us}}(p)$, let 
$\hat{y}^\star(p,g)=\arg\max_{\hat{y}\in\hat{Y}(p)} \mathrm{lcp}(I_g,I_{\hat{y}})$
(if $\hat{Y}(p)=\varnothing$, define $\mathrm{lcp}=0$ and $\hat{y}^\star$ undefined).
Let $\pi(p,g)$ be the shared prefix of $I_g$ and $I_{\hat{y}^\star}$ (use the empty prefix $\epsilon$ when $\mathrm{lcp}=0$).
Let $\mathcal{H}$ denote the \emph{hierarchical label tree} used to define indices, and $
\mathcal{C}_{\mathcal{H}}(\pi)=\{\,x \mid \pi \preceq I_x\,\}$ be the set of concepts whose index starts with prefix $\pi$ in $\mathcal{H}$. Define the \emph{number of ontology concepts} as $T \triangleq |\mathcal{C}_{\mathcal{H}}(\epsilon)|$.
For each $g$, set $S(p,g)=|\mathcal{C}_{\mathcal{H}}(\pi(p,g))|$, with the convention $S(p,g)=T$ when $\mathrm{lcp}=0$.
We aggregate sizes via a harmonic mean:
\[
\mathrm{U\text{-}CS}
\;=\;
\Bigg(
\frac{1}{\sum_{p}|G_{\text{us}}(p)|}
\sum_{p}\;\sum_{g\in G_{\text{us}}(p)} \frac{1}{S(p,g)}
\Bigg)^{-1}.
\]
Lower is better; using the harmonic mean reduces sensitivity to rare extremely large candidate sets.

% \paragraph{Example (toy).}
% If $I_g=[0,2,5,2,0]$ and predictions include $[0,7,8,7,8,0]$ and $[0,2,5,8,2]$, then $\mathrm{lcp}$ lengths are $1$ and $3$, so the per-item closeness is $3/5=0.6$ and $\pi(g)=[0,2,5]$; $\mathrm{U\text{-}CS}$ counts the concepts under that subtree.

\medskip
\noindent
Together, micro-F1, U-RC, and U-CS capture whether models \textit{hit} the target, \textit{approach} it in the hierarchy, and \textit{shrink} the search space of it.

\section{Results}

\subsection{RQ1: Is LLM-generated ALD sufficient for training effective recognizers?}
% \subsection{RQ1: Is LLM-generated ALD sufficient for training a recognizer that performs effectively on both seen and unseen concepts?}
\begin{table}[t]
    \centering
    \small
    \begin{tabular}{l|ccc}
    \toprule
    \multicolumn{4}{c}{HPA-MLD (228 instances)}\\
    \midrule
        \textbf{Pipeline} & \textbf{Pre} & \textbf{Rec} & \textbf{F1}\\
    \midrule
        PCC & 53.0 & \textbf{53.0} & 53.0 \\
        $\rightarrow$ Concept Classification (CC) & 68.6 & 46.9 & 55.7$^{\dagger}$\\
        $\rightarrow$ Relabeling (RL) & 66.3 & 49.3 & \textbf{56.6}$^{\dagger}$\\
        $\rightarrow$ QS & 54.1 & 50.8 & 52.4$\star$\\
        $\rightarrow$ CC $\rightarrow$ RL & 70.7 & 44.9 & 54.9$^{\dagger}$ \\ 
        $\rightarrow$ CC $\rightarrow$ RL $\rightarrow$ QS & \textbf{71.3} & 45.4 & 55.4$^{\dagger}$\\
    \toprule
    \multicolumn{4}{c}{HoIP-MLD (500 instances)}\\
    \midrule
    %     \textbf{Pipeline} & \textbf{Pre} & \textbf{Rec} & \textbf{F1}\\
    % \midrule
        PCC& 6.3 & 62.4 & 11.4\\
        $\rightarrow$ Concept Classification (CC) & 12.0 & 51.7 & 19.5$^{\dagger}$\\
        $\rightarrow$ Relabeling (RL) & 8.0 & 50.6 & 13.8$^{\dagger}$ \\
        $\rightarrow$ GF & 11.1 & 44.0 & 17.7$^{\dagger}$ \\
        $\rightarrow$ QS & 7.2 & \textbf{65.0} & 12.9${\star}$\\
        $\rightarrow$ CC $\rightarrow$ RL & 14.3 & 43.3 & 21.5$^{\dagger}$\\
        $\rightarrow$ CC $\rightarrow$ RL $\rightarrow$ GF & \textbf{20.3} & 40.1 & 27.0$^{\dagger}$\\
        $\rightarrow$ CC $\rightarrow$ RL $\rightarrow$ GF $\rightarrow$ QS & 20.2 & 43.6 & \textbf{27.6}$^{\dagger}$ \\

    \bottomrule
    \end{tabular}
    \caption{
    Auto-labeling performance on manually labeled datasets (MLDs). ``PCC,'' ``GF,'' and ``QS'' denote the stages of Passage-to-Claim-to-Concept generation, Guideline-based Filtering, and Quality-based Selection, respectively. A $^{\dagger}$ indicates a significant difference between the annotations produced by ``PCC'' and those of a given pipeline, as determined by a one-sided paired t-test ($p < 0.05$). A $^{\star}$ indicates one-sided paired t-test is not applicable, because the remaining instances after ``PCC-QS'' share same annotations with ``PCC.''  %The ``Score'' column reports the average quality ratings assigned by the LLM-based scoring prompt in the QS stage.
    }
    \label{tab:ald-ori-hpo}
\end{table}

\begin{table*}[t]
\small
% \footnotesize
% \scriptsize
    \centering
    \begin{tabular}{c|ccc|ccc|ccc}
\toprule
\multicolumn{10}{c}{HoIP-MLD}\\
\midrule
\multirow{2}{*}{\textbf{\(|D|\)}}& \multicolumn{3}{c|}{\textbf{OSI}}& \multicolumn{3}{c|}{\textbf{SSI}}&\multicolumn{3}{c}{\textbf{OSSI}}\\
% \cmidrule{2-16}
& \textbf{F1} &  \textbf{U-RC} & \textbf{U-CS$\downarrow$} & \textbf{F1} &  \textbf{U-RC} & \textbf{U-CS$\downarrow$} & \textbf{F1} & \textbf{U-RC} & \textbf{U-CS$\downarrow$}\\
\midrule
200 & 38.5  & 27.4 & 303.8 & 40.1 & 22.0 & 215.8 & 44.0 & \textbf{27.8} & \textbf{139.8} \\
400 & 50.9  & 26.7 & \textbf{84.1} & 52.8  & 24.3 & 110.6 & \textbf{63.9}  & \textbf{27.9} & 228.6 \\
800 & 59.4 & 28.0 & \textbf{51.4} & 74.5 & 26.1 & 70.2 & \textbf{67.3} & \textbf{30.0} & 136.5 \\
1,600 & 70.9 & 30.2 & 82.8 & 71.1  & 29.0 & \textbf{71.5} & \textbf{73.7}  & \textcolor{red}{\textbf{31.5}} & 271.8 \\
2,458 & 73.1 & 29.7 & 67.4 & 77.3 & 28.4 & \textcolor{red}{\textbf{42.1}} & \textcolor{red}{\textbf{78.6}} & \textbf{30.5} & 150.8 \\
\toprule
\multicolumn{10}{c}{HoIP-ALD}\\
% \midrule
% \multirow{2}{*}{\textbf{\(|D|\)}}& \multicolumn{3}{c|}{\textbf{OSI}}& \multicolumn{3}{c|}{\textbf{SSI}}&\multicolumn{3}{c}{\textbf{OSSI}}\\
% \cmidrule{2-16}
% & \textbf{F1} &  \textbf{U-RC} & \textbf{U-CS$\downarrow$} & \textbf{F1} &  \textbf{U-RC} & \textbf{U-CS$\downarrow$} & \textbf{F1} & \textbf{U-RC} & \textbf{U-CS$\downarrow$}\\
\midrule
200 & \textcolor{red}{\textbf{18.5}}  & \textbf{30.1} & 211.7 & 13.7  & 28.6 & \textbf{129.1} & 16.8 & 28.5 & 313.0 \\
400 & 13.9 & \textbf{32.6} & 243.0 & \textbf{14.1} & 28.6 & \textbf{95.0} & 13.4 & 32.4 & 120.2 \\
800 & \textbf{14.9} & \textbf{35.4} & 172.9 & 13.9 & 31.9 & \textbf{75.9} & 11.9 & 34.7 & 103.7 \\
1,600 & 15.9 & \textbf{38.0} & 109.7 & 14.8 & 36.0 & \textbf{55.0} & \textbf{18.4} & 35.1 & 94.2 \\
3,200 & \textbf{18.3} & 37.5 & 106.5 & 15.0 & 36.1 & \textbf{44.8} & 15.5 & \textbf{39.3} & 62.5 \\
6,400 & \textbf{17.2} & \textbf{40.8} & 69.6 & 15.7 & 38.4 & \textbf{35.0} & 15.9 & 40.6 & 52.1 \\
12,800 & \textbf{16.9} & \textbf{41.1} & 67.5 & 16.0 & 38.7 & \textbf{34.6} & 16.7 & 40.7 & 51.5 \\
16,415 & \textbf{17.7} & 40.9 & 73.3 & 17.6 & 38.0 & \textcolor{red}{\textbf{33.2}} & 17.3 & \textcolor{red}{\textbf{41.4}} & 44.4\\
\toprule
\multicolumn{10}{c}{HPA-ALD}\\
% \midrule
% \multirow{2}{*}{\textbf{\(|D|\)}}& \multicolumn{3}{c|}{\textbf{OSI}}& \multicolumn{3}{c|}{\textbf{SSI}}&\multicolumn{3}{c}{\textbf{OSSI}}\\

% & \textbf{F1} &  \textbf{U-RC} & \textbf{U-CS$\downarrow$} & \textbf{F1} &  \textbf{U-RC} & \textbf{U-CS$\downarrow$} & \textbf{F1} & \textbf{U-RC} & \textbf{U-CS$\downarrow$}\\
% \cmidrule{2-16}

\midrule
200 & 19.1 & 26.5 & 100.6 & \textbf{20.4} & \textbf{27.0} & \textbf{76.0} & 16.7 & 26.8 & 103.0 \\
400 & \textbf{20.3} & 28.2 & 68.1 & 17.4  & 29.5 & 64.0 & 19.2 & \textbf{30.2} & \textbf{61.7} \\
800 & \textbf{20.6} & \textbf{30.6} & \textbf{55.0} & 19.2 & 30.2 & 60.2 & 17.7 & 30.5 & 55.1 \\
1,600 & \textbf{20.4} & 33.7 & 70.3 & 20.3 & 33.4 & 59.4 & 19.1 & \textbf{34.5} & \textbf{41.4} \\
3,200 & \textbf{25.7} & \textbf{39.3} & 39.7 & 22.9 & 35.7 & 45.6 & 20.7 & 36.9 & \textbf{37.2} \\
6,400 & \textbf{26.0} & \textbf{41.7} & \textbf{32.3} & 25.0 & 38.2 & 45.3 & 23.6 & 39.3 & 35.0 \\
12,800 & \textbf{28.6} & 39.4 & 41.0 & 26.8 & 38.6 & 36.4 & 28.2 & \textbf{40.1} & \textbf{29.3} \\
25,600 & \textbf{34.9} & 43.4 & 35.5 & 33.4 & 41.2 & \textbf{24.7} & 32.4 & \textbf{46.0} & 25.7 \\
47,152 & 35.3  & \textcolor{red}{\textbf{46.1}} & \textcolor{red}{\textbf{20.1}} & 36.0 & 41.2 & 24.7 & \textcolor{red}{\textbf{36.7}} & 45.6 & 20.5 \\
\bottomrule
    \end{tabular}
    \caption{Performance on HoIP concept recognition using models trained on HoIP-MLD, and performance on HoIP and HPA concept recognition using models trained on HoIP-ALD and HPA-ALD. $|D|$ denotes training data size. F1 denotes Micro-F1. U-RC refer to Unseen Recall-oriented Closeness. U-CS indicates Unseen Candidate set Size. The best score across indices is highlighted in bold, while the best score achieved by a type of training data is in red. }
    \label{tab:HoIP-ALD-res}
\end{table*}

To answer the RQ1, we first evaluate the performance of our annotation pipeline and then assess the performance of models trained on ALD.

\paragraph{Annotation Pipeline Effectiveness.}
As shown in Table~\ref{tab:ald-ori-hpo}, each module in our pipeline contributes to the final annotation quality. Concept classification, relabeling, and guideline-based filtering consistently improve precision by refining candidates, while quality-based selection enhances recall. When all modules are combined, the pipeline achieves strong gains in both Precision (53.0$\rightarrow$71.3 for HPA; 6.3$\rightarrow$20.2 for HoIP) and F1 scores (11.4$\rightarrow$27.6 for HoIP), validating its design. 
However, labels generated by our LLM-based pipeline are significantly misaligned with manual annotations (F1 scores are quite low). This gap is particularly pronounced for the novel and fine-grained HoIP ontology, whose concepts are poorly represented in current LLMs, leading to low data annotation quality. Overall, errors introduced by LLM generation and similarity-based matching can propagate across stages, and later filtering and relabeling only partially mitigate them. As a result, residual noise and passage-level quality variation remain, motivating systematic evaluation of auto-labeled data quality and downstream usefulness.

% Without the QS stage, our pipeline can be viewed as an unsupervised approach to MA-BCR, which shows still limited performance. The limitation is particularly stark for HoIP (F1: 27.0), a more complex and recently developed ontology with limited prior NLP research.

% Removing quality-based selection yields a fully unsupervised recognition pipeline, whose limited performance, especially for HoIP (e.g., HoIP F1 maxes out at 27.6), highlights the challenges in scaling to new ontologies. HoIP concepts are semantically fine-grained and newly opened (2024), lacking representation in current LLMs. This results in high conceptual ambiguity and low precision even after refinement. 

\paragraph{Recognizers Trained Solely on ALD.}
We proceed to train the MA-COIR using only the generated ALD. The results, presented in Table~\ref{tab:HoIP-ALD-res}, indicate that models trained on HoIP-ALD underperform their MLD-trained counterparts in terms of overall F1 score. 
For instance, even the best ALD-trained model achieves an F1 of only 18.5, which is substantially lower than the worst supervised baseline trained on just 200 manually labeled abstracts (Using OSI, 200 instances in HoIP-MLD for training, F1 is 38.5). 
The significant differences between ALD- and MLD-trained models highlights that \textbf{LLM-generated ALD is not a replacement for high-quality manual annotations at present}.

\subsection{RQ2: Does ALD Improve Generalization to Unseen Concepts Despite Noise?}

Given that ALD cannot match the performance of MLD for recognizer training, we investigate its complementary value with our evaluation framework.
%, particularly  for the generalization to unseen concepts. 
Our results reveal a key insight: while annotation quality constrains exact-match performance, \textbf{models can learn structural information even from imperfect annotations.} As shown in Table~\ref{tab:HoIP-ALD-res}, when the training data volume of HoIP-ALD increases, models show consistent improvement on metrics specifically designed to measure generalization \textbf{regardless of index types}.\footnote{As comparisons between indices are less relevant to our research questions, we provide the analysis in Append~\ref{app:com-index}.} % (U-RC$\uparrow$, U-CS$\downarrow$). 
Additionally, OSSI achieves the best U-RC, while SSI achieves the best U-CS on HoIP-ALD. This demonstrates that the two metrics are not redundant and each capturing a unique and important facet of generalization. 
The evaluation framework is model-agnostic and can be applied to any recognizer; for models without hierarchical indices, concepts can be mapped to OSI/SSI/OSSI before evaluation. 

As ALD volume increases, the improvements in U-RC confirm that predictions get closer to the unseen gold concepts within the hierarchy. %, even when an exact match is not achieved. 
U-CS shrinks substantially—from 129.1 to 33.2 on HoIP using SSI, and down to 20.1 on HPA using OSI. This demonstrates that the model, trained with more ALD, learns to effectively ruling out implausible concepts from the entire label set. This trend is even more pronounced on HPA, where higher-quality ALD enables all metrics, including F1, to improve with data volume. These results stand in clear contrast to MLD-trained models, where we observed no clear correlation between training data size and unseen concept recognition, likely due to limited ontology coverage (0.53–1.75\%). 

% We identify two primary reasons that ALD enhances generalization: (1) ALD dramatically increases the number and diversity of concepts the model is exposed to, providing the necessary scale and structural context for learning to generalize. (2) Hierarchical indices provide a structured learning framework that helps the model make sense of the noisy ALD supervision. As demonstrated in prior work, there is a significant performance gap between non-hierarchical and hierarchical indices \citep{liu2025macoirleveragingsemanticsearch}. The performance differences achieved by OSI/SSI/OSSI on HoIP-MLD also highlight the importance of indices. Our results confirm that \textbf{hierarchical indexing and ALDs are complementary}: indices structure the noisy supervision, while ALDs expand the coverage and reinforce the hierarchical signals. 

We attribute ALD’s generalization gains to two factors: (1) ALD substantially expands concept coverage and diversity, providing scale and structural context for learning. (2) Hierarchical indices impose a structure that makes noisy ALD supervision learnable. Consistent with prior findings \citep{liu2025macoirleveragingsemanticsearch}, and with the OSI/SSI/OSSI gaps observed on HoIP-MLD (combining ontological and semantic signals brings the best F1), indexing plays a critical role. Overall, \textbf{hierarchical indexing and ALDs are complementary}: indices structure noise, while ALDs broaden coverage and reinforce hierarchical signals.

\section{Additional Analyses}
This section provides complementary analyses to better interpret our results. We first validate the downstream utility of U-RC and U-CS in a recognition-to-reranking setting, and then analyze false positive patterns in the LLM-based auto-labeling pipeline to clarify its limitations.

\subsection{Downstream Utility}

\begin{table}[]
\small
    \centering
    \begin{tabular}{c|ccc|ccc}
\toprule
\multirow{2}{*}{\textbf{\(|D|\)}}&  \multicolumn{3}{c|}{\textbf{Recognizer}} & \multicolumn{3}{c}{\textbf{Reranker}} \\
 & \textbf{F1}  &  \textbf{U-RC}  &  \textbf{U-CS}   &  \textbf{P}  &  \textbf{R}  & \textbf{F1} \\
\midrule
  200 & 14.7 & 25.4 & 118.9 & 41.6 & 29.4 & 33.7\\
 400 & 14.3 & 27.8 & 127.6 & 35.5 & 29.4 & 32.2\\
 800 & 14.7 & 28.9 & 77.1 & 41.8 & 29.0 & 34.2\\
 1,600 & 13.8 & 33.9 & 88.3 & \textbf{42.7} & 30.5 & 35.6\\
 3,200 & 16.2 & 32.6 & 95.1 & 40.2 & 31.1 & 35.1\\
 6,400 & 15.0 & 34.9 & 59.3 & 40.1 & 33.5 & 36.5\\
 12,800 & 15.7 & \textbf{38.0} & \textbf{41.4} & 37.3 & 34.0 & 35.6\\
 16,415 & \textbf{17.2} & 36.7 & 45.3 & 40.1 & \textbf{39.6} & \textbf{39.8}\\
\bottomrule
    \end{tabular}
    \caption{Recognition-to-reranking results under a fixed retrieval budget.}
    \label{tab:down-res}
\end{table}

\begin{figure}
    \centering
    \includegraphics[width=1.0\linewidth]{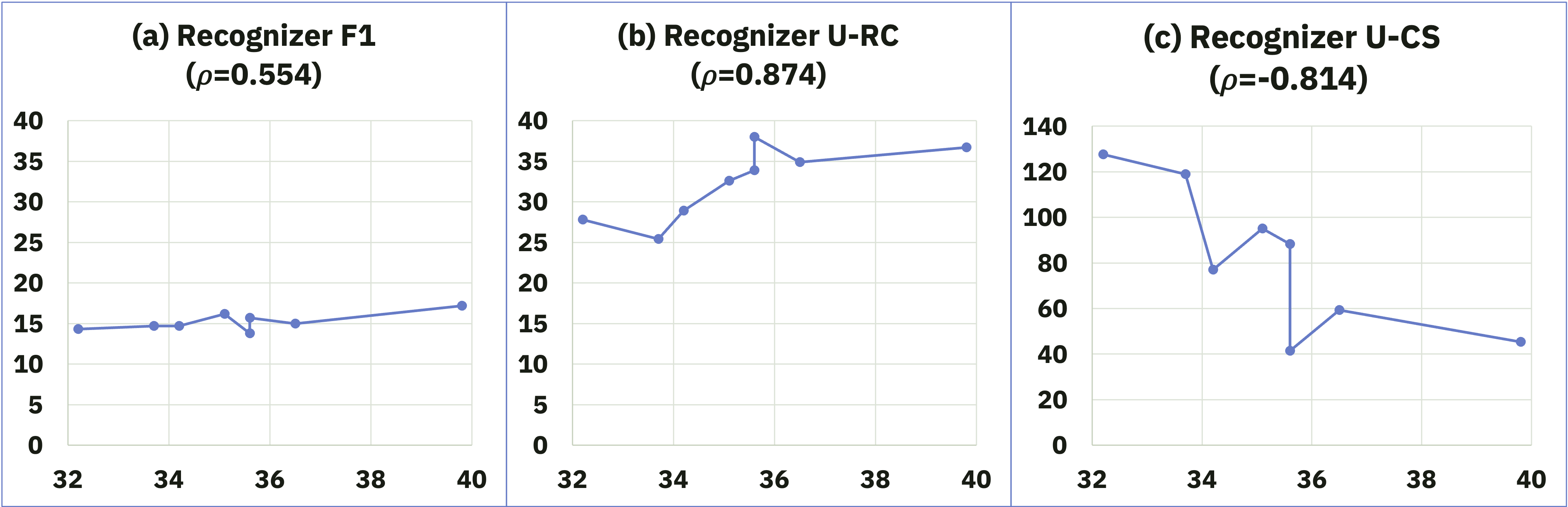}
    \caption{Upstream metrics vs. downstream reranker F1. Each panel plots one recognizer metric against the reranker F1 across eight recognizers trained with different data volumes. $\rho$ denotes Spearman’s rank correlation between the upstream metric and reranker F1.}
    \label{fig:down-res}
\end{figure}

To validate the downstream utility of U-RC and U-CS, we conduct a controlled recognition-to-reranking experiment on MLD-HoIP using predictions from eight ALD-trained recognizers (using SSI) over all test passages. For each predicted concept, we retrieve 41 candidate concepts (as the smallest U-CS among recognizers) from the ontology using BM25 and merge the retrieved candidates within each passage to form a passage-level candidate pool. This pool is then reranked by an MLD-trained cross-encoder ($|D| = 2,458$).\footnote{Details are provided in Appendix~\ref{app:reranker}.}

Table~\ref{tab:down-res} reports downstream performance under this fixed retrieval budget. Reranker F1 scores vary from 32.2 to 39.8, even when the recognizer micro-F1 remains relatively flat across several training sizes (e.g., 13.8–16.2 from 200 to 3,200), indicating that exact-match accuracy alone does not fully reflect downstream usefulness. Figure~\ref{fig:down-res} further shows that downstream reranker F1 aligns most strongly with U-RC ($\rho{=}0.874$) and negatively with U-CS ($\rho{=}{-}0.814$), whereas recognizer micro-F1 exhibits weaker rank correlation ($\rho{=}0.554$). Together, these results suggest that U-RC and U-CS capture downstream-relevant quality signals beyond exact-match F1 performance.

\subsection{False Positive Error Analysis}
\label{sec:fp-analysis}

\begin{figure}
    \centering
    \includegraphics[width=1.0\linewidth]{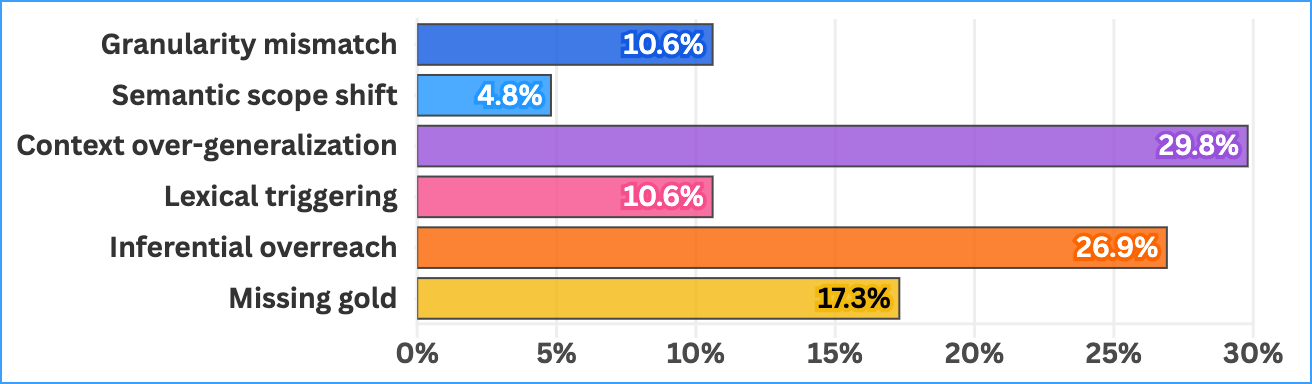}
    \caption{Distribution of false positive outcomes in the LLM-based auto-labeling pipeline. ``Missing gold'' denotes predictions supported by the passage but absent from dataset-provided gold annotations; the remaining bars correspond to five error types.}
    \label{fig:fp-dist}

\end{figure}

We manually analyzed false positive predictions (FPs) from the LLM-based auto-labeling pipeline using a small random sample from HoIP-MLD. Specifically, we randomly selected 20 instances from the 500-instance subset for auto-labeling evaluation, covering 37 gold concepts and 104 initially labeled false positives. We filtered out FPs that are passage-supported but missing from the dataset-provided annotations, as ``Missing gold''; 17.3\% of apparent FPs fall into this category (see Figure~\ref{fig:fp-dist}). The remaining FPs are grouped into five error types: \textit{granularity mismatch, semantic scope shift, context over-generalization, lexical triggering, and inferential overreach} (definitions in Appendix~\ref{app:error-taxonomy}).

Among true errors, \textit{context over-generalization} (29.8\%) and \textit{inferential overreach} (26.9\%) dominate among true errors, indicating that LLMs often expand beyond passage evidence by turning partial cues or outcomes into broader process claims. The remaining errors include \textit{lexical triggering} (10.6\%), \textit{granularity mismatch} (10.6\%), and \textit{semantic scope shift} (4.8\%). Overall, this analysis clarifies structural limitations of LLM-based auto-labeling—particularly its tendency to over-interpret passage evidence.

% \section{Conclusion}

% We evaluate generalization in mention-agnostic biomedical concept recognition (CR) with an unseen-aware setup and two hierarchy-aware metrics. To our knowledge, we introduce the first LLM-based auto-labeling pipeline for this setting and construct two large auto-labeled datasets (ALDs), rigorously assessing their quality and usefulness. Comparing ALD-trained and Manual-Labeled-Data-trained models reveals a consistent trade-off: ALD lowers F1 yet improves unseen generalization, indicating that broad, noisy supervision contributes coverage and structural signal. Our setup, metrics, pipeline, and datasets provide a reproducible basis for measuring, and ultimately improving performance on unseen CR.

\section{Conclusion}
This work advances Mention-agnostic Biomedical Concept Recognition (MA-BCR) in two key directions: (1) We propose an evaluation framework with hierarchical indices and unseen-aware metrics to quantify generalization. (2) To the best of our knowledge, we introduce the first LLM-based auto-labeling pipeline for MA-BCR and demonstrate that corresponding ALD, while low-quality, enhances generalization to unseen concepts by supplementing coverage and structural information. %Our setup, metrics, pipeline, and datasets provide a reproducible basis for measuring, and ultimately improving performance on MA-BCR.

\newpage

\section*{Limitations}

Despite promising results, several limitations warrant further attention:

\textbf{Auto-labeling limitations on complex ontologies.} While our LLM-based auto-labeling pipeline substantially outperforms direct concept generation, its performance on HoIP remains limited. This reflects the difficulty LLMs face when annotating novel, fine-grained biomedical concepts they probably have never seen during pretraining. Further improvements are needed to enhance precision and recall in underrepresented subdomains. Additionally, using a single LLM across multiple stages (stage 2-5) of auto-labeling may not fully realize the potential of the LLM; different LLMs may perform better at different stages and mitigate error propagation between stages.

\textbf{Incomplete concept coverage.} Our auto-labeled datasets currently cover only 70\% of HPA and 54\% of HoIP concepts. Many ontology terms fail to retrieve any relevant PubMed abstracts, limiting data diversity. Expanding retrieval strategies and a new data synthesis strategy may help improve both coverage and representativeness.

\textbf{Indexing strategy remains improvable.} OSSI already improves generalization by integrating structural and semantic signals, but more sophisticated methods—e.g., adaptive edge weighting, concept importance modeling, or learned fusion mechanisms—may further enhance performance, especially under noisy supervision.

\textbf{Scalability of training.} As labeled concept coverage grows, training on larger datasets using encoder-decoder models like BART becomes time-intensive. While ontologies are relatively stable, improving training efficiency (e.g., via distillation, continual learning, or lightweight architectures) will be key to scaling this approach in real-world applications.

\textbf{Automatic evaluation reliability.} Finally, although our use of GPT-4o-mini enables scalable evaluation (as the Quality-based selection stage), we found LLM-as-judge still does not perform ideal alignment with manually crafted standards. Human validation remains necessary to calibrate automatic metrics and ensure robust assessment.

\section*{Acknowledgments}
We thank all reviewers for their careful reading and constructive feedback. We are particularly grateful to the meta-reviewer for their dedicated effort in coordinating the review discussion and providing an exceptionally detailed and clear meta-review. Their thorough reading and thoughtful guidance significantly strengthened this work.

\bibliography{acl_latex}

@inproceedings{tan-etal-2024-large,
    title = "Large Language Models for Data Annotation and Synthesis: A Survey",
    author = "Tan, Zhen  and
      Li, Dawei  and
      Wang, Song  and
      Beigi, Alimohammad  and
      Jiang, Bohan  and
      Bhattacharjee, Amrita  and
      Karami, Mansooreh  and
      Li, Jundong  and
      Cheng, Lu  and
      Liu, Huan",
    editor = "Al-Onaizan, Yaser  and
      Bansal, Mohit  and
      Chen, Yun-Nung",
    booktitle = "Proceedings of the 2024 Conference on Empirical Methods in Natural Language Processing",
    month = nov,
    year = "2024",
    address = "Miami, Florida, USA",
    publisher = "Association for Computational Linguistics",
    url = "https://aclanthology.org/2024.emnlp-main.54/",
    doi = "10.18653/v1/2024.emnlp-main.54",
    pages = "930--957"
}

@inproceedings{de2011generalized,
  title={Generalized louvain method for community detection in large networks},
  author={De Meo, Pasquale and Ferrara, Emilio and Fiumara, Giacomo and Provetti, Alessandro},
  booktitle={2011 11th international conference on intelligent systems design and applications},
  pages={88--93},
  year={2011},
  organization={IEEE}
}

@article{karypis1997metis,
  title={METIS: A software package for partitioning unstructured graphs, partitioning meshes, and computing fill-reducing orderings of sparse matrices},
  author={Karypis, George and Kumar, Vipin},
  year={1997}
}

@article{zhang2021fast,
  title={Fast multi-resolution transformer fine-tuning for extreme multi-label text classification},
  author={Zhang, Jiong and Chang, Wei-Cheng and Yu, Hsiang-Fu and Dhillon, Inderjit},
  journal={Advances in Neural Information Processing Systems},
  volume={34},
  pages={7267--7280},
  year={2021}
}

@article{kharbanda2022cascadexml,
  title={Cascadexml: Rethinking transformers for end-to-end multi-resolution training in extreme multi-label classification},
  author={Kharbanda, Siddhant and Banerjee, Atmadeep and Schultheis, Erik and Babbar, Rohit},
  journal={Advances in neural information processing systems},
  volume={35},
  pages={2074--2087},
  year={2022}
}

@misc{tay2022transformermemorydifferentiablesearch,
      title={Transformer Memory as a Differentiable Search Index}, 
      author={Yi Tay and Vinh Q. Tran and Mostafa Dehghani and Jianmo Ni and Dara Bahri and Harsh Mehta and Zhen Qin and Kai Hui and Zhe Zhao and Jai Gupta and Tal Schuster and William W. Cohen and Donald Metzler},
      year={2022},
      eprint={2202.06991},
      archivePrefix={arXiv},
      primaryClass={cs.CL},
      url={https://arxiv.org/abs/2202.06991}, 
}

@artical{Yamagata2024,
    year ={2024},
    title={Homeostasis imbalance process ontology: a study on COVID-19 infectious processes},
    author ={Yuki Yamagata and Tatsuya Kushida and Shuichi Onami and Hiroshi Masuya},
    jounral ={BMC Medical Informatics and Decision Making},
    url ={https://doi.org/10.1186/s12911-024-02516-0},
    doi ={10.1186/s12911-024-02516-0},
}

@misc{wang2023exploringincontextlearningability,
      title={Exploring the In-context Learning Ability of Large Language Model for Biomedical Concept Linking}, 
      author={Qinyong Wang and Zhenxiang Gao and Rong Xu},
      year={2023},
      eprint={2307.01137},
      archivePrefix={arXiv},
      primaryClass={cs.CL},
      url={https://arxiv.org/abs/2307.01137}, 
}

@article{luo2021phenotagger,
  title={PhenoTagger: a hybrid method for phenotype concept recognition using human phenotype ontology},
  author={Luo, Ling and Yan, Shankai and Lai, Po-Ting and Veltri, Daniel and Oler, Andrew and Xirasagar, Sandhya and Ghosh, Rajarshi and Similuk, Morgan and Robinson, Peter N and Lu, Zhiyong},
  journal={Bioinformatics},
  volume={37},
  number={13},
  pages={1884--1890},
  year={2021},
  publisher={Oxford University Press}
}

@article{DBLP:journals/biodb/LiSJSWLDMWL16,
  author    = {Jiao Li and
               Yueping Sun and
               Robin J. Johnson and
               Daniela Sciaky and
               Chih{-}Hsuan Wei and
               Robert Leaman and
               Allan Peter Davis and
               Carolyn J. Mattingly and
               Thomas C. Wiegers and
               Zhiyong Lu},
  title     = {BioCreative {V} {CDR} task corpus: a resource for chemical disease
               relation extraction},
  journal   = {Database J. Biol. Databases Curation},
  volume    = {2016},
  year      = {2016},
  url       = {https://doi.org/10.1093/database/baw068},
  doi       = {10.1093/database/baw068},
  bibsource = {dblp computer science bibliography, https://dblp.org}
}

@article{10.1093/bioinformatics/btae104,
    author = {Caufield, J Harry and Hegde, Harshad and Emonet, Vincent and Harris, Nomi L and Joachimiak, Marcin P and Matentzoglu, Nicolas and Kim, HyeongSik and Moxon, Sierra and Reese, Justin T and Haendel, Melissa A and Robinson, Peter N and Mungall, Christopher J},
    title = "{Structured Prompt Interrogation and Recursive Extraction of Semantics (SPIRES): a method for populating knowledge bases using zero-shot learning}",
    journal = {Bioinformatics},
    volume = {40},
    number = {3},
    pages = {btae104},
    year = {2024},
    month = {02},
    issn = {1367-4811},
    doi = {10.1093/bioinformatics/btae104},
    url = {https://doi.org/10.1093/bioinformatics/btae104},
    eprint = {https://academic.oup.com/bioinformatics/article-pdf/40/3/btae104/56912793/btae104.pdf},
}

@article{10.1093/nar/gkad1005,
    author = {Gargano, Michael A  et al.},
    title = {The Human Phenotype Ontology in 2024: phenotypes around the world},
    journal = {Nucleic Acids Research},
    volume = {52},
    number = {D1},
    pages = {D1333-D1346},
    year = {2023},
    month = {11},
    issn = {0305-1048},
    doi = {10.1093/nar/gkad1005},
    url = {https://doi.org/10.1093/nar/gkad1005},
    eprint = {https://academic.oup.com/nar/article-pdf/52/D1/D1333/55039688/gkad1005.pdf},
}

@inproceedings{el-khettari-etal-2024-mention,
    title = "Mention-Agnostic Information Extraction for Ontological Annotation of Biomedical Articles",
    author = "El Khettari, Oumaima  and
      Nishida, Noriki  and
      Liu, Shanshan  and
      Munne, Rumana Ferdous  and
      Yamagata, Yuki  and
      Quiniou, Solen  and
      Chaffron, Samuel  and
      Matsumoto, Yuji",
    editor = "Demner-Fushman, Dina  and
      Ananiadou, Sophia  and
      Miwa, Makoto  and
      Roberts, Kirk  and
      Tsujii, Junichi",
    booktitle = "Proceedings of the 23rd Workshop on Biomedical Natural Language Processing",
    month = aug,
    year = "2024",
    address = "Bangkok, Thailand",
    publisher = "Association for Computational Linguistics",
    url = "https://aclanthology.org/2024.bionlp-1.37",
    pages = "457--473",
}

@article{groza2024evaluation,
  title={An evaluation of GPT models for phenotype concept recognition},
  author={Groza, Tudor and Caufield, Harry and Gration, Dylan and Baynam, Gareth and Haendel, Melissa A and Robinson, Peter N and Mungall, Christopher J and Reese, Justin T},
  journal={BMC Medical Informatics and Decision Making},
  volume={24},
  number={1},
  pages={30},
  year={2024},
  publisher={Springer}
}

@article{douze2024faiss,
      title={The Faiss library},
      author={Matthijs Douze and Alexandr Guzhva and Chengqi Deng and Jeff Johnson and Gergely Szilvasy and Pierre-Emmanuel Mazaré and Maria Lomeli and Lucas Hosseini and Hervé Jégou},
      year={2024},
      eprint={2401.08281},
      archivePrefix={arXiv},
      primaryClass={cs.LG}
}

@misc{liu2025macoirleveragingsemanticsearch,
      title={MA-COIR: Leveraging Semantic Search Index and Generative Models for Ontology-Driven Biomedical Concept Recognition}, 
      author={Shanshan Liu and Noriki Nishida and Rumana Ferdous Munne and Narumi Tokunaga and Yuki Yamagata and Kouji Kozaki and Yuji Matsumoto},
      year={2025},
      eprint={2505.12964},
      archivePrefix={arXiv},
      primaryClass={cs.CL},
      url={https://arxiv.org/abs/2505.12964}, 
}

@article{https://doi.org/10.1155/2017/8565739,
author = {Lobo, Manuel and Lamurias, Andre and Couto, Francisco M.},
title = {Identifying Human Phenotype Terms by Combining Machine Learning and Validation Rules},
journal = {BioMed Research International},
volume = {2017},
number = {1},
pages = {8565739},
doi = {https://doi.org/10.1155/2017/8565739},
url = {https://onlinelibrary.wiley.com/doi/abs/10.1155/2017/8565739},
eprint = {https://onlinelibrary.wiley.com/doi/pdf/10.1155/2017/8565739},
year = {2017}
}

\appendix

\section{Appendix}

\subsection{Auto-Labeled Data Construction}
\label{app:alp}

\begin{figure*}[t]
  \includegraphics[width=\linewidth]{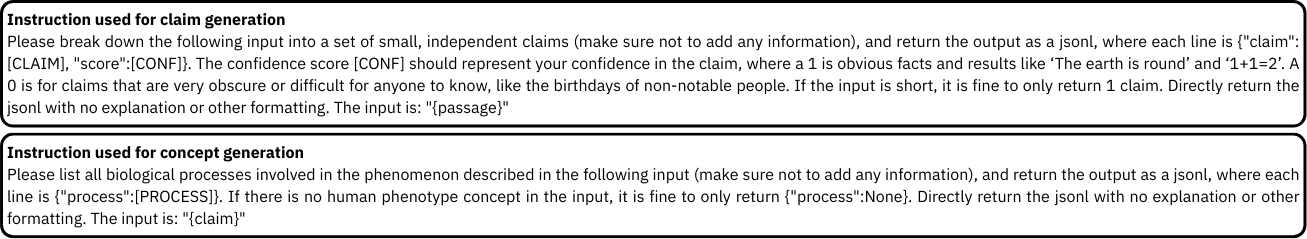}
  \caption{The prompts we used for PCC stage for HoIP concept generation.}
  \label{fig:prompt-pcc}
\end{figure*}

\begin{figure*}[t]
  \includegraphics[width=\linewidth]{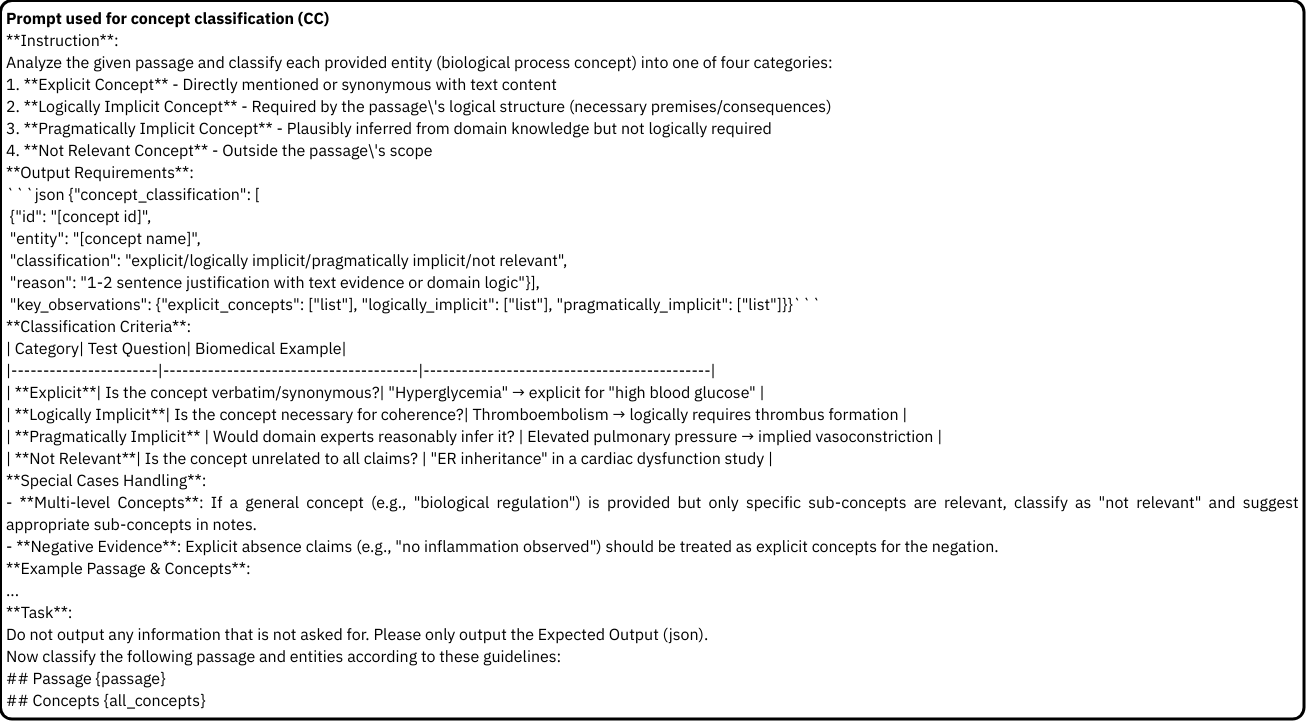}
  \caption{The prompt used for concept classification.}
  \label{fig:prompt-cc}
\end{figure*}

\begin{figure*}[t]
  \includegraphics[width=\linewidth]{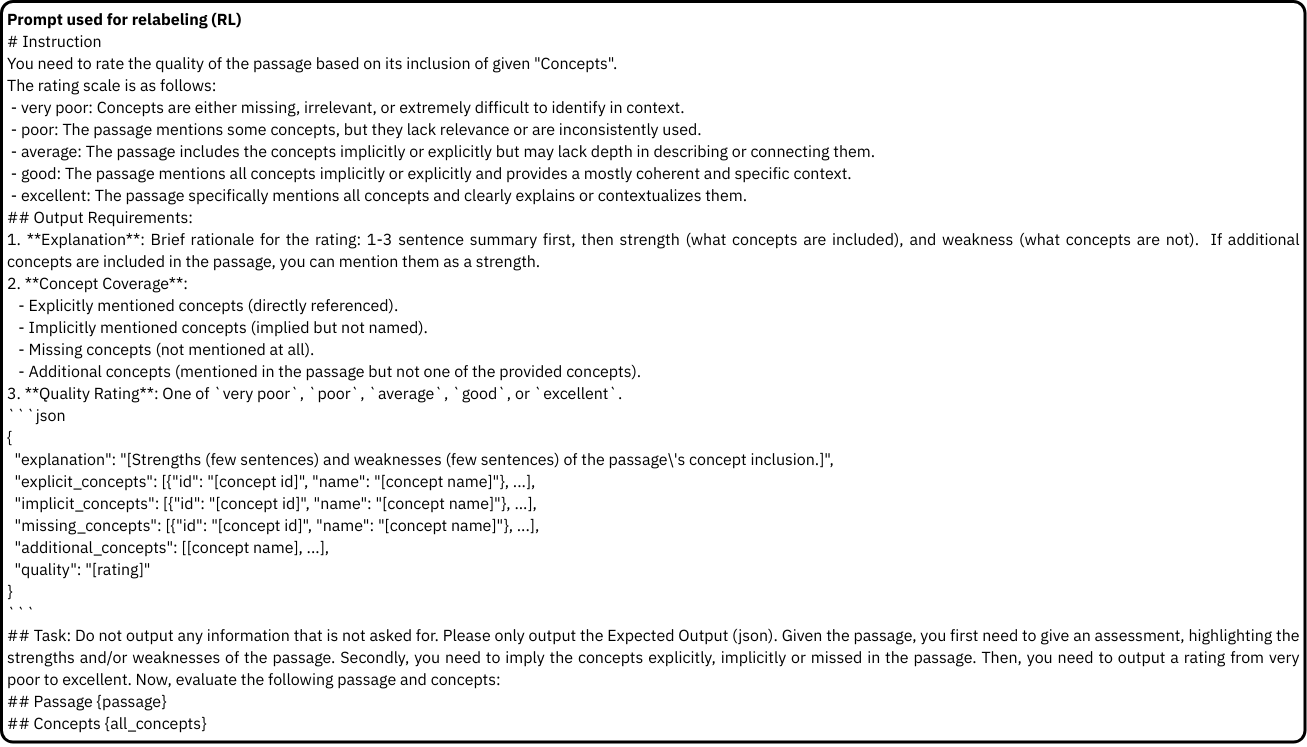}
  \caption{The prompt used for relabeling.}
  \label{fig:prompt-rl}
\end{figure*}

\begin{figure*}[t]
  \includegraphics[width=\linewidth]{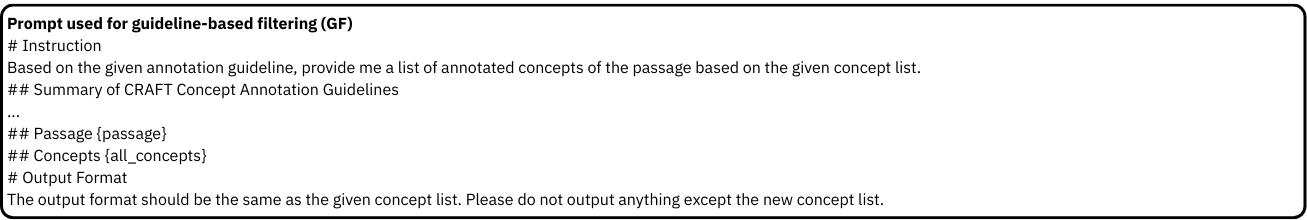}
  \caption{The prompt used for guideline-based filtering.}
  \label{fig:prompt-gf}
\end{figure*}

\subsubsection{Collect passages for annotation}

We retrieved PubMed article abstracts using ontology term names as search keywords. For each term, a single representative name was used to query PubMed via the NIH E-utilities API\footnote{https://www.ncbi.nlm.nih.gov/books/NBK25500/}, requesting up to the 10 most recent articles. The abstracts of the retrieved articles were collected as "passages". Although we aimed to obtain 10 abstracts per concept, some terms yielded fewer or no results. Multiple terms could be associated with the same articles, we removed duplicate abstracts to ensure all passages in the dataset are unique.

\subsubsection{Prompts}

Prompts of Passage-to-Claim-to-Concept generation, Concept Classification, Relabeling, and Guideline-based Filtering are listed in Figure~\ref{fig:prompt-pcc}, ~\ref{fig:prompt-cc}, ~\ref{fig:prompt-rl}, ~\ref{fig:prompt-gf}, respectively.

In Quality-based Selection stage, we use the prompt similar as the relabeling step (Figure~\ref{fig:prompt-rl}), but remove the output request for ``explicit/implicit/missing/additional concept'', and only take the ``Quality'' returned by the LLMs for processing.

\subsubsection{Inference costs}

While the pipeline may appear complex, it was designed to replace expensive manual annotation for biomedical ontologies like HPO and HoIP, which lack high-coverage existing training corpora. For instance, annotating 300 HoIP abstracts manually took nearly 6 months by a domain expert. In contrast, our pipeline processes an abstract in under 60 seconds, with a total cost under \$150 to generate 90,000+ annotated abstracts—orders of magnitude cheaper and faster.
We deliberately selected affordable LLMs (e.g., Llama-3-8B, GPT-4o-mini) over more costly alternatives. We have found Llama-3-8B achieves better scores than GPT-3.5-turbo, GPT-4o-mini and GPT-4o in Passage-to-Claim-to-Concept generation stage. We have found GPT-4o improves F1 by only 0.5 but increases cost 16× in the Concept Classification stage. We believe our design strikes a practical balance between scalability and performance. 

We implement Llama-3-8B on one NVIDIA A100 Tensor Core GPU, so there is no API cost for Passage-to-Claim-to-Concept generation, and the processing time is average 22.8 seconds per abstract.

The cost of each stage, given the candidate list obtained by Passage-to-Claim-to-Concept generation using GPT-4o-mini for annotating HPA concepts for one abstract in HPO GSC+ corpus, is listed in the Table~\ref{tab:cost} for reference.
The whole pipeline costs less than simply taking the cost in the table as a total, because each stage will remove part of the concepts, so that the input and output tokens will be reduced.

As Guideline-based Filtering stage only is applied for HoIP concept annotation, we report the cost of annotating 100 passages from CRAFT corpus using GPT-4o-mini, given the candidate lists obtained by Passage-to-Claim-to-Concept generation for reference: 701 seconds and 0.04 dollar.

\begin{table}[t]
\small
    \centering
    \begin{tabular}{c|cc}
    \hline
    Stage & Time (s)& API cost (\$)\\
    \hline
         Concept Classification& 6.81 & 0.0003\\
         Relabeling & 5.62 & 0.0003\\
         Quality-based Selection & 4.04 & 0.0003 \\
    \hline
    \end{tabular}
    \caption{Costs for annotating one abstract in the CRAFT corpus using GPT-4o-mini. The candidate lists obtained by Passage-to-Claim-to-Concept generation are used as inputs. The number is an average among 100 abstracts.}
    \label{tab:cost}
\end{table}

\subsection{Current limitations of MLDs.}
\label{app:dataset}
Accurate evaluation of BCR systems relies on high-quality, manually annotated text-concept pairs. In this work, we focus on two critical categories: Human Phenotype Abnormality (HPA), documented in the Human Phenotype Ontology (HPO)\footnote{https://hpo.jax.org/}, and Homeostasis Imbalance Process (HoIP), documented in the Homeostasis Imbalance Process Ontology (HOIP)\footnote{https://bioportal.bioontology.org/ontologies/HOIP}. While the HPO GSC+ dataset provides reliable annotations for HPA, manually curated resources for HoIP remain scarce and inconsistent, posing significant challenges for systematic evaluation. To partially address this gap, we repurposed the CRAFT corpus—a dataset annotated with Gene Ontology (GO)\footnote{http://purl.obolibrary.org/obo/go.owl} terms—leveraging the conceptual overlap between HoIP and GO. For HPA, we used HPO GSC+ annotations restricted to descendants of ``HP:0000118-Phenotypic abnormality''. For HoIP, we used CRAFT passages annotated with GO terms subsumed by the HoIP ontology.

\begin{figure*}[t]
  \includegraphics[width=\linewidth]{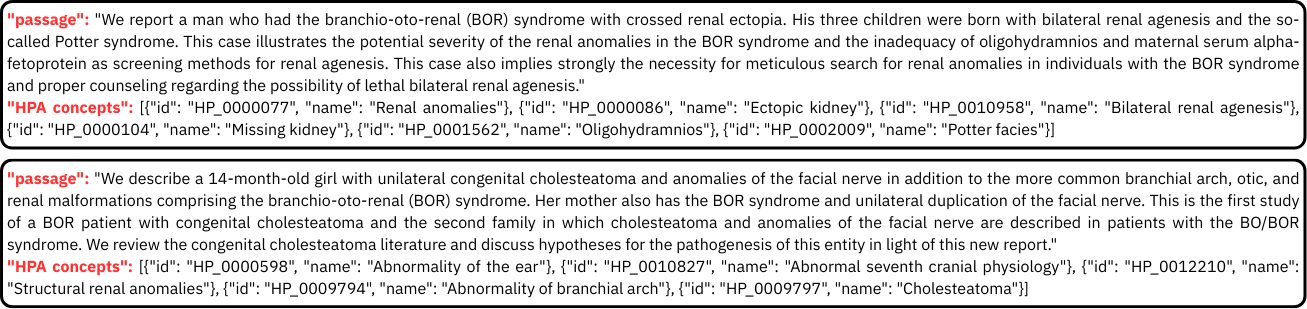}
  \caption{Two instances in HPA-MLD (from HPO GSC+ corpus).}
  \label{fig:example-hpa}
\end{figure*}

\begin{figure*}[t]
  \includegraphics[width=\linewidth]{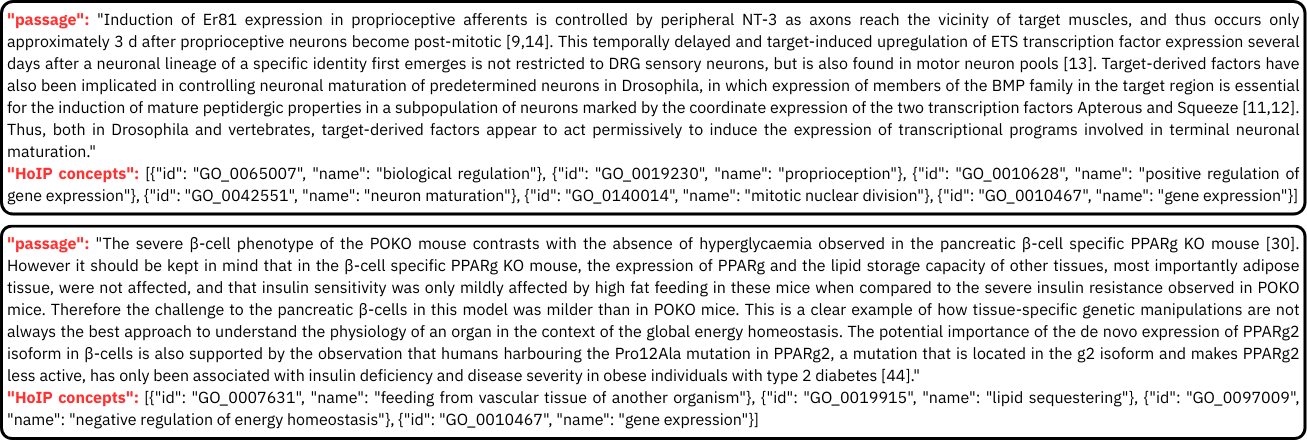}
  \caption{Two instances in HoIP-MLD (from CRAFT corpus).}
  \label{fig:example-hoip}
\end{figure*}

We show some passage-concept pairs as examples in Figure \ref{fig:example-hpa} and Figure \ref{fig:example-hoip} for better understanding what kind of concepts we are targeted.

\paragraph{Inherent challenges in manual annotation}
Our analysis uncovered inconsistencies in annotation practices. In the HPO GSC+ corpus, fine-grained terms (e.g., bilateral vestibular schwannoma) are often annotated alongside their hypernyms (vestibular schwannoma, schwannoma), but this hierarchy is not applied systematically. For example, broader parent terms are occasionally omitted even when their descendants are labeled, reflecting inconsistencies in annotation protocols.

The CRAFT corpus presents even more pronounced limitations. Annotated in 2010, it lacks contemporary GO refinements: some terms are now obsolete (e.g., former Homeostasis Imbalance Processes reclassified as molecular functions), while some are annotated at overly broad levels by current standards. Although we excluded obsolete terms, correcting overly general annotations to match modern ontologies would require infeasible manual effort. Consequently, our evaluation necessarily inherits these historical biases.

\paragraph{The scarcity problem.}
As Table~\ref{tab:data-stats}-MLD illustrates, the available manual annotations cover only 2.35\% of target ontology concepts by both HPO GSC+ and CRAFT. This extreme sparsity—coupled with the inconsistencies described above—underscores a key bottleneck: manual annotations are insufficient to robustly evaluate CR systems, let alone train them.

\paragraph{The ratio of implicit concepts.}
No existing dataset explicitly distinguishes implicit vs. explicit concepts in annotations, so a ratio cannot be reliably calculated. According to our own observation, there is around 20\% of annotated concepts implicit with respect to their corresponding annotated mentions.

% \paragraph{Implications for methodology.}
% These findings motivate our subsequent use of automatically labeled datasets to supplement scarce human annotations. While we employ existing MLDs for performance validation, our results will be interpreted with explicit consideration of their flaws: limited coverage, inconsistent granularity, and temporal obsolescence. 

\subsection{Details of Concept Recognizer}
\label{app:model}
\paragraph{Hyperparameters.}
The BART-based language model (facebook/bart-large) used in MA-COIR for recognition is trained with hyperparameters listed in the Table \ref{tab:hyp}. We trained all models on one NVIDIA A100 Tensor Core GPU. For one training instance, the experiment ran for an average of 1.82s/it.

\begin{table}[t]
\small
    \centering
    \begin{tabular}{c|c}
    \hline
    Item & Value \\
    \hline
         model\_card &  facebook/bart-large\\
         learning\_rate& 1e-5\\
         num\_epoch& 50\\
         batch\_size& 4\\
         max\_length\_of\_tokens & 1024 \\
    \hline
    \end{tabular}
    \caption{Hyperparameters of the recognizer.}
    \label{tab:hyp}
\end{table}

\paragraph{Data splits.}
The statistics of each data split used for concept recognizer training are shown in the Table~\ref{tab:split-stats}.

\subsection{Details of Hierarchical Graph Partitioning}
\label{app:gp-algo}
We apply Louvain clustering by default \cite{de2011generalized}, and switch to METIS \cite{karypis1997metis} when subgraphs are too sparse or chain-like to meet the constraint ($\leq 10$ children of each internal node). 

\begin{figure}[t]
    \centering
    \includegraphics[width=1.0\linewidth]{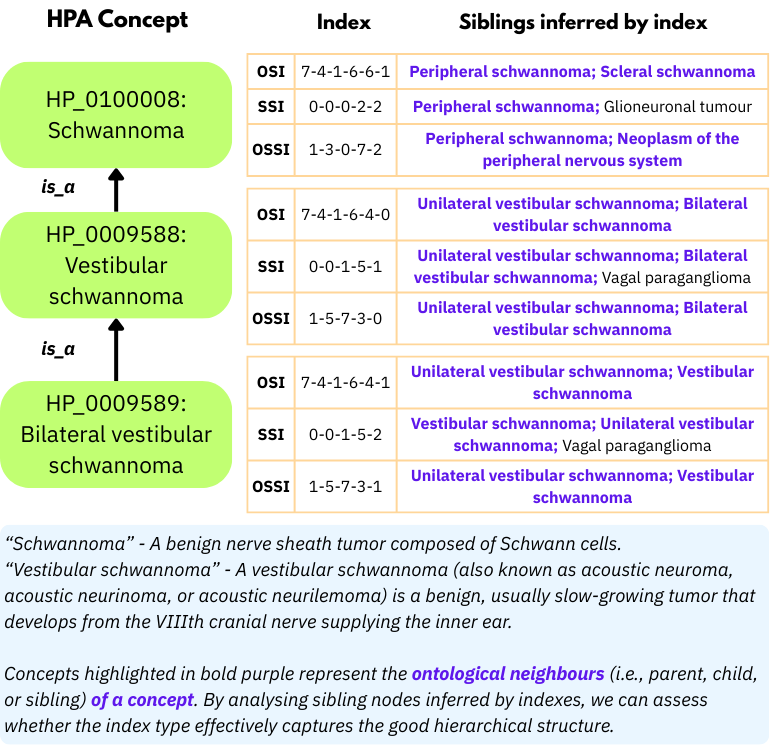}
    \caption{An example of constructed search indices. When the information used for graph construction is modified, a concept may share the same index prefix with different concepts.}
    \label{fig:hpo-index}
\end{figure}

An example of search indices is shown in Figure~\ref{fig:hpo-index}.

\subsection{Comparison of Hierarchical indices}
\label{app:com-index}
As shown in Table~\ref{tab:HoIP-ALD-res}, On HoIP, OSSI achieves the highest U-RC, while HPA consistently favors OSI. SSI most effectively reduces the candidate set size of unseen concepts on HoIP (lowest U-CS), whereas OSI and OSSI are comparable on HPA. These differences derive from the structural properties of the ontologies: HPO is a shallow, relatively balanced tree, allowing OSI to be both discriminative and learnable. In contrast, HoIP Ontology is deep and entangled, with multi-parent nodes and root-to-leaf paths that may exceed 10 layers. In this setting, OSI produces longer index sequences (avg. depth: 6.15 vs. 5.42 for SSI) and suffers from branching imbalance. While the lack of ontological information degrades F1, SSI compensates by narrowing candidate spaces more aggressively. OSSI balances both structural and semantic signals, yielding superior overall performance. 

Additionally, OSSI achieves the best U-RC, while SSI achieves the best U-CS on HoIP-ALD. This demonstrates that the two metrics are not redundant and each capturing a unique and important facet of generalization. U-RC measures hierarchical proximity but not the reduction in search space. A prediction and the gold concept may reside in the same penultimate cluster that contains 2 or more candidates—a critical difference U-RC misses. Therefore, we propose the U-CS. We highlight the importance of \textbf{selecting evaluation metrics aligned with downstream application goals}. When predictions are used directly, closeness metrics like F1 and U-RC are paramount. However, when downstream modules perform precision refinement, candidate space reduction (e.g., U-CS) becomes the more relevant indicator.

\begin{table*}[t]
\small
% \scriptsize
\centering
    \begin{tabular}{ccc|ccc}
    \toprule
\multicolumn{6}{c}{HoIP-MLD}\\
\midrule
\multicolumn{3}{c|}{Training}& \multicolumn{3}{c}{Test} \\
\midrule
\(|D|\)& \(| Concept |\)& Coverage (\%)& \(| Concept |\)& \(| Unseen\_Concept |\)& Seen (\%)\\
\midrule
200& 155& 0.53& 229& 227& 38.86\\
400& 236& 0.80& 229& 160& 53.71\\
800& 313& 1.07& 229& 144& 58.52\\
1,600& 423& 1.44& 229& 119& 66.81\\
2,458& 515& 1.75& 229& 111& 69.00\\
\toprule
\multicolumn{6}{c}{HoIP-ALD}\\
\midrule
% \multicolumn{3}{c|}{Training}& \multicolumn{3}{c}{Test} \\
% \midrule
% \(|D|\)& \(| Concept |\)& Coverage(\%)& \(| Concept |\)& \(| Unseen\_Concept |\)& Seen (\%)\\
% \midrule
200& 744& 2.53& 635& 477& 24.88\\
400& 1,400& 4.77& 635& 424& 33.23\\
800& 2,399& 8.17& 635& 379& 40.31\\
1,600& 3,881& 13.22& 635& 328& 48.35\\
3,200& 5,801& 19.75& 635& 286& 54.96\\
6,400& 8,181& 27.86& 635& 246& 61.26\\
12,800& 10,844& 36.93& 635& 226& 64.41\\
16,415& 11,848& 40.34& 635& 219& 65.51\\
    \toprule
\multicolumn{6}{c}{HPA-ALD}\\
% \midrule
% \multicolumn{3}{c|}{Training}& \multicolumn{3}{c}{Test} \\
% \midrule
% \(|D|\)& \(| Concept |\)& Coverage(\%)& \(| Concept |\)& \(| Unseen\_Concept |\)& Seen (\%)\\
\midrule
200& 416& 2.27& 386& 304& 21.24\\
400& 848& 4.62& 386& 288& 25.39\\
800& 1,523& 8.30& 386& 264& 31.61\\
1,600& 2,515& 13.70& 386& 234& 39.38\\
3,200& 3,942& 21.48& 386& 207& 46.37\\
6,400& 5,724& 31.19& 386& 186& 51.81\\
12,800& 7,918& 43.14& 386& 163& 57.77\\
25,600& 10,188& 55.51& 386& 137& 64.51\\
47,152& 11,989& 65.32& 386& 125& 67.62\\
\bottomrule

    \end{tabular}
        \caption{Dataset statistics of splits for recognizer training and evaluation. $|D|$ denotes training data size. $| Concept |$ denotes the number of unique concepts. ``Coverage(\%)'' refers to the percentage of ontology concepts presented in the dataset. ``Seen(\%)'' refers to the percentage of test concepts presented in the training set.}
    \label{tab:split-stats}
\end{table*}

%—whether under low-quality HoIP-ALD or high-quality HPA-ALD—particularly for unseen concepts.

\subsection{Details of Concept Reranker}
\label{app:reranker}

\paragraph{Model.}
Our downstream reranker is a cross-encoder implemented with the HuggingFace Transformers library using \texttt{AutoModelForSequenceClassification} (\texttt{num\_labels}=1), initialized from a SapBERT checkpoint. Given a passage and a candidate concept text, the model outputs a scalar relevance score for reranking.

\paragraph{Training objective and negatives.}
We train the reranker with a listwise softmax loss over groups consisting of one positive concept and $k$ hard negatives ($k{=}4$ by default). For each training passage--gold concept pair, we use the gold concept text as a BM25 query to retrieve a candidate list from the ontology and treat the retrieved non-gold concepts as hard negatives (20 per passage--gold concept pair). Hard negatives are randomly resampled per batch to improve robustness.

\paragraph{Model selection on gold-based candidates.}
For dev-time model selection, we evaluate using the same supervision format as training: for each passage--gold concept pair, we construct a BM25 candidate set using the gold concept text and measure reranking quality (e.g., nDCG@K/MRR@K). Early stopping is applied based on this gold-based dev evaluation, ensuring that model selection does not depend on recognizer-generated candidates.

\paragraph{Final evaluation on recognizer-based candidates.}
For the downstream utility experiment, candidate sets are generated from recognizer predictions: each predicted concept retrieves a fixed number of ontology candidates via BM25 (41 per prediction), which are merged (deduplicated union) into a passage-level pool and reranked. We select a score threshold on the dev split by maximizing micro-F1, and then apply the dev-selected threshold to compute micro Precision/Recall/F1 on the test split.

\subsection{Error taxonomy}
\label{app:error-taxonomy}

To enable systematic characterization of false positive predictions in biological process annotation, we define an error taxonomy consisting of five categories.

Predictions are evaluated against a passage-supported reference set constructed as follows: we begin with the dataset-provided gold annotations, and additionally include biological process concepts that are explicitly stated or unambiguously implied in the passage but missing from the original gold. 
These missing concepts are identified through manual review of predictions initially labeled as false positives. 
Accordingly, a prediction is considered erroneous only if it cannot be justified by passage evidence under this reference set. 
This refinement step corrects for incomplete coverage in the original dataset annotations and does not depend on model predictions beyond using them to surface candidate missing concepts for review.

Importantly, error categories describe \textbf{how} an unsupported prediction arises, rather than simply whether it is incorrect. 

\subsubsection{Definition of error types}

\paragraph{E-1: Granularity Mismatch}
A granularity mismatch occurs when the passage supports a specific concept, but the prediction selects a term that is hierarchically related in the ontology (an ancestor or descendant node) while failing to match the precise level of abstraction evidenced in the text. These errors are strictly confined to the same ontology lineage; the model identifies the correct conceptual path but fails to calibrate the node's depth relative to the textual evidence.

\paragraph{E-2: Semantic Scope Shift}
This category applies when a prediction deviates from the semantic domain of the target concept to an unrelated class. Unlike the hierarchical depth errors in E-1, a scope shift represents a thematic misalignment where the prediction belongs to an entirely different conceptual branch. For instance, misidentifying a \textit{learning} event as a \textit{memory} event changes the underlying biological phenomenon being documented, thereby misrepresenting the specific functional activity characterized in the passage.

\paragraph{E-3: Context Over-generalization} This error arises when the presence of an isolated observation or a specific biological outcome is misinterpreted as sufficient evidence for a broader, encompassing concept that is not itself stated. In these cases, the textual evidence may describe a necessary component or a resultant state of a process, but the existence of the process as a whole is not grounded in the passage. Unlike the hierarchical positioning in E-1, E-3 represents an inductive leap where partial information is erroneously treated as the occurrence of a complex, multi-stage event.

\paragraph{E-4: Lexical Triggering} Lexical triggering refers to predictions driven by the presence of salient keywords or surface patterns rather than by the actual conceptual information. These errors reflect a reliance on stereotypical associations, where the mention of a domain-specific term---such as the name of a gene or protein (\textit{e.g., "Shh"}) or a specific body part (\textit{e.g., "testis"})---biases the model toward predicting an associated concept (\textit{e.g., "gene expression"} or \textit{e.g., "male gonad development"}) that is not explicitly supported by the text. Here, the model's internal statistical bias regarding the keyword overrides the contextual evidence.

\paragraph{E-5: Inferential Overreach} Inferential overreach captures predictions that necessitate unstated logical steps or external assumptions beyond the explicit description. While E-3 over-generalizes from an observed outcome to a broader conceptual class, E-5 adds entirely new explanatory logic that is absent from the text. This occurs when the model presupposes missing intermediate connections, moving from a simple descriptive statement to an unsupported causal chain.

\subsubsection{Diagnostic Criteria}
To ensure consistent classification, we distinguish the five categories via a three-step decision logic:
\begin{enumerate}
    \item \emph{Lineage Calibration:} If the prediction and a supported concept share the same ontological path (i.e., one is an ancestor of the other), is the error solely due to the node's depth? (If yes $\rightarrow$ \emph{E-1})
    \item \emph{Domain Alignment:} Does the prediction shift to a fundamentally different semantic class or an unrelated branch within the ontology? (If yes $\rightarrow$ \emph{E-2})
    \item \emph{Evidence Gap Analysis:} Is the lack of support due to over-extending a partial observation (\emph{E-3}), keyword-driven bias (\emph{E-4}), or the addition of an unsupported causal chain (\emph{E-5})?
\end{enumerate}

\end{document}